\documentclass{article}

\PassOptionsToPackage{numbers, compress}{natbib}
 \usepackage[preprint]{neurips_2026}

\usepackage[utf8]{inputenc}
\usepackage[T1]{fontenc}
\usepackage{hyperref}
\usepackage{url}
\usepackage{microtype}
\usepackage{url}
\usepackage{booktabs}
\usepackage{times}
\usepackage{latexsym}

\usepackage[T1]{fontenc}

\usepackage[utf8]{inputenc}

\usepackage{microtype}

\usepackage{inconsolata}
\usepackage{booktabs}
\usepackage{amsfonts}
\usepackage{amsmath}
\usepackage{amssymb}
\usepackage{microtype}
\usepackage{graphicx}
\graphicspath{{figures/}}
\usepackage{xcolor}
\usepackage{subcaption}
\usepackage{enumitem}
\usepackage{tabularx}
\usepackage{multirow}
\usepackage{colortbl}
\usepackage{tikz}
\usepackage{booktabs}
\usepackage{amsfonts}
\usepackage{amsmath}
\usepackage{amssymb}
\usepackage{microtype}
\usepackage{graphicx}
\usepackage{xcolor}
\usepackage{subcaption}
\usepackage{tabularx}
\usepackage{array}
\usepackage{multirow}
\usepackage{enumitem}
\usepackage{ragged2e} 
\newcolumntype{L}{>{\RaggedRight\arraybackslash}X} 
\setlist[itemize]{nosep,leftmargin=*}
\usepackage{hyperref}
\usepackage{url}
\usepackage{tcolorbox}
\usepackage{listings}
\usepackage{multirow}
\usepackage{enumitem}
\usepackage{xcolor}
\usepackage{algorithm}
\usepackage{algpseudocode}

\usepackage{etoolbox}
\makeatletter
\newcommand{\@Cref@getprefix}[1]{\@Cref@split#1:\@nil}
\def\@Cref@split#1:#2\@nil{#1}
\newcommand{\Cref}[1]{%
  \edef\@Cref@pfx{\@Cref@getprefix{#1}}%
  \def\@Cref@word{}%
  \ifdefstring{\@Cref@pfx}{fig}{\def\@Cref@word{Figure}}{}%
  \ifdefstring{\@Cref@pfx}{tab}{\def\@Cref@word{Table}}{}%
  \ifdefstring{\@Cref@pfx}{sec}{\def\@Cref@word{Section}}{}%
  \ifdefstring{\@Cref@pfx}{app}{\def\@Cref@word{Appendix}}{}%
  \ifdefstring{\@Cref@pfx}{eq}{\def\@Cref@word{Equation}}{}%
  \@Cref@word~\ref{#1}%
}
\makeatother
\usepackage{lineno}
\newcommand{\bca}{\textsc{BCA}}
\newcommand{\ctg}{\textsc{CTG}}
\newcommand{\dcc}{\textsc{Detect-Classify-Compare}}
\usepackage{microtype}               
\usepackage[skip=4pt]{caption}       
\title{When Reasoning Traces Become Performative:\\ Step-Level Evidence that Chain-of-Thought Is an Imperfect Oversight Channel}

\author{
\textbf{Wenkai Li}$^{1,2}$,
\textbf{Fan Yang}$^{2}$,
\textbf{Shaunak A. Mehta}$^{2}$,
\textbf{Ananya Hazarika}$^{2}$,
\textbf{Koichi Onoue}$^{2}$\\
$^{1}$Carnegie Mellon University \quad
$^{2}$Fujitsu Research of America Inc.\\
\texttt{\{wenkail\}@cs.cmu.edu}
}

\begin{document}
\maketitle

\begin{abstract}
Chain-of-thought (CoT) traces are increasingly used both to improve language-model capability and to audit model behavior, implicitly assuming that the visible trace remains synchronized with the computation that determines the answer. We test this assumption with a step-level \emph{Detect-Classify-Compare} framework built around an answer-commitment proxy that is cross-validated with Patchscopes, tuned-lens probes, and causal direction ablation. Across nine models and seven reasoning benchmarks, latent commitment and explicit answer arrival align on only 61.9\% of steps on average. The dominant mismatch pattern is \emph{confabulated continuation}: 58.0\% of detected mismatch events occur after the answer-commitment proxy has already stabilized while the trace continues producing deliberative-looking text, and a vacuousness analysis shows that the committed answer does not change during these steps. In architecture-matched Qwen2.5/DeepSeek-R1-Distill comparisons, the reasoning pipeline changes failure composition more than aggregate alignment, most clearly at 32B where confabulated steps decrease as contradictory states increase. Lower step-level alignment is also associated with larger CoT utility, suggesting that the settings that benefit most from CoT are often the least temporally faithful. Paired truncation and a complementary donor-corruption test further indicate that much pure-CS text is not load-bearing for the final answer. These findings suggest that CoT can remain useful while still being an unreliable report of when the answer was formed.
\end{abstract}

\section{Introduction}
\label{sec:intro}

Chain-of-thought (CoT) reasoning began as a prompting technique for improving multi-step performance~\citep{wei2022chain,kojima2022large}, but in current reasoning model pipelines it also serves as a training target, a verification interface, and an oversight signal~\citep{lightman2023verify,wang2024mathshepherd,deepseek2025r1,openai2024o1,baker2025cot_monitorability,kenton2025cot_evasion}. Humans and monitor models increasingly inspect reasoning traces to judge whether a system reasoned safely, honestly, or competently. This dual role rests on a strong assumption: that the visible trace faithfully tracks the internal computation that produced the answer. If that assumption fails, CoT may still improve capability while providing a misleading basis for oversight.

A growing literature suggests that this assumption is fragile. Perturbation, bias-injection, and deployment studies document omitted hints, post-hoc rationalization, and other forms of unfaithful CoT~\citep{lanham2023measuring,turpin2024language,paul2024making,chen2025reasoning,arcuschin2025chain,li2026personanongratasinglemethod}. Conceptual and empirical work further suggests that faithfully verbalizing autoregressive computation may itself be difficult~\citep{creswell2022faithful,tanneru2024hardness,saparov2023language,nye2021work}. The question is therefore no longer whether unfaithful CoT exists, but how often it arises in current reasoning models and what kind of mismatch it creates.

Yet the field still lacks a measurement lens tailored to this oversight question. Most existing faithfulness methods deliver response-level verdicts~\citep{lanham2023measuring,paul2024making,shen2026faithcotbench}: they tell us whether a trace is suspect, but not when the answer was already internally fixed relative to the visible chain. Concurrent work by \citet{boppana2026reasoning_theater} studies a closely related phenomenon under the label \emph{performative CoT}, showing that reasoning traces can continue after latent answer commitment, especially on easier tasks. Related work by \citet{palod2025performative_thinking} questions whether longer CoTs should be interpreted as evidence of more adaptive computation. Our focus is complementary but narrower. We center the \emph{timing} of internal answer formation, distinguish five step-level mismatch patterns rather than a binary split, compare architecture-matched Qwen2.5/R1 pairs across three scales, and quantify the relationship between temporal faithfulness and CoT utility. Interpretability tools can decode emerging predictions inside the model~\citep{nostalgebraist2020logitlens,belrose2023eliciting,ghandeharioun2024patchscopes,marks2023geometry}, but they have rarely been used to measure CoT faithfulness at the step level.

\begin{figure*}[t]
\centering
\includegraphics[width=1\textwidth]{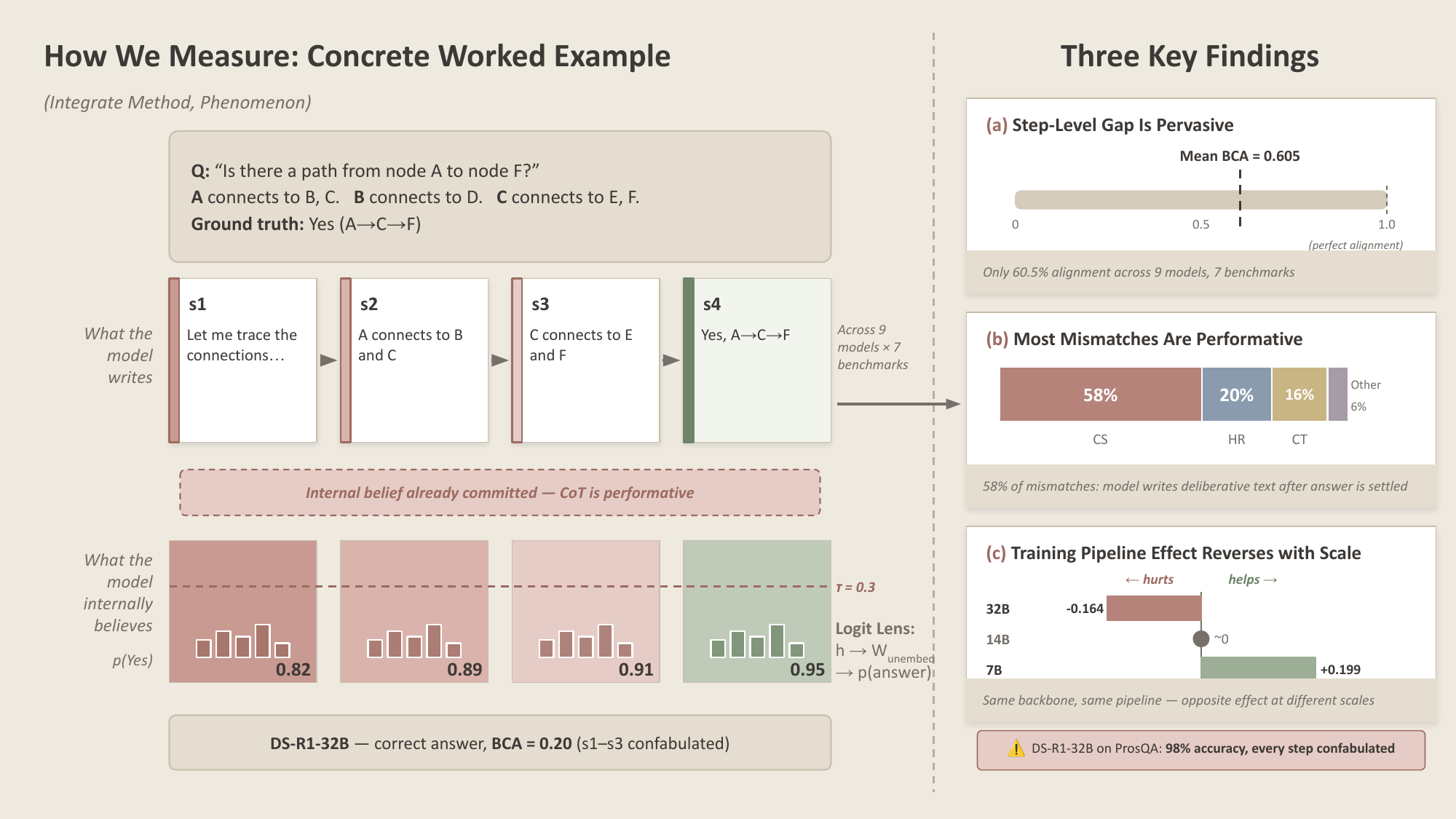}
\caption{Overview of the empirical picture. Left: on a ProsQA example, DS-R1-32B commits to the answer by step~1 while the written trace continues, yielding low \bca{}. Right: the aggregate results show the same pattern at scale, with 61.9\% mean alignment, 58\% of mismatches classified as confabulated steps, and pipeline effects that vary across model scales.}
\label{fig:overview}
\end{figure*}

We introduce the \dcc{} framework to close this gap. At each CoT step, we extract an answer-commitment proxy using logit lens projections, cross-validate that proxy with Patchscopes concordance, tuned-lens probes, and direction ablation, then quantify timing divergence, classify mismatch events, and compare matched model pairs that differ in whether they underwent the DeepSeek-R1 distillation pipeline. Across nine models and seven reasoning benchmarks (\Cref{fig:overview}), the dominant failure mode is not overt contradiction but \emph{post-commitment continuation}: the model often resolves the answer internally and then continues generating plausible reasoning text. The resulting picture is one in which a fluent and even accuracy-improving chain can still be a poor report of when the answer was formed.

Our investigation makes three contributions:
\begin{enumerate}[leftmargin=*,itemsep=2pt,topsep=2pt]
    \item We introduce a step-level framework for measuring whether internal answer formation is temporally reflected in CoT, and find mean alignment of only 61.9\%, with 58.0\% of mismatch events falling into \emph{confabulated} continuation. 
    \item We show that in architecture-controlled Qwen2.5/DeepSeek-R1-Distill comparisons, the reasoning pipeline changes not only aggregate alignment but also the composition of failures, most sharply at 32B where confabulated steps fall as contradictory states rise.
    \item We connect this timing measure to downstream behavior: lower step-level alignment tends to coincide with larger CoT utility, while paired truncation and donor corruption indicate that much post-commitment text is descriptive rather than load-bearing.
    \end{enumerate}
    
\section{Related Work}
\label{sec:related}

\paragraph{CoT faithfulness, oversight, and monitoring.}
Chain-of-thought prompting improves reasoning performance~\citep{wei2022chain,kojima2022large,yao2024tree,nye2021work,wang2024chainofthought}, but whether the resulting trace is a faithful explanation or oversight signal remains contested~\citep{wiegreffe2021teach,deyoung2020eraser}. Perturbation, bias-injection, and deployment studies show that models can preserve answers under trace edits, omit decisive information, or produce post-hoc rationalizations~\citep{lanham2023measuring,turpin2024language,chen2025reasoning,arcuschin2025chain}. This matters because CoT is increasingly used in process supervision, verifier training, and safety monitoring workflows, where monitorability may fail even when CoT remains useful for capability~\citep{lightman2023verify,wang2024mathshepherd,baker2025cot_monitorability,kenton2025cot_evasion,macdermott2025reasoning_under_pressure,zolkowski2025obfuscate,meek2025monitorability}. Most of this literature provides response-level verdicts~\citep{radhakrishnan2023question,roger2023benchmarking,shen2026faithcotbench}; our question is narrower and more temporal: when does internal answer formation diverge from the visible trace?

\paragraph{Internal readouts and performative CoT.}
Interpretability tools such as the logit lens~\citep{nostalgebraist2020logitlens}, tuned lens~\citep{belrose2023eliciting}, Patchscopes~\citep{ghandeharioun2024patchscopes}, and linear probes~\citep{marks2023geometry,li2024inference,burns2023discovering,azaria2023internal,kadavath2022language} make it possible to track emerging predictions during generation. We use these tools not to recover a full reasoning circuit, but to estimate when answer commitment becomes decodable. Closest to our setting, \citet{boppana2026reasoning_theater} study performative CoT and show that latent answer commitment can precede what the trace reveals, especially on easier tasks. \citet{palod2025performative_thinking} similarly cautions against interpreting longer CoTs as evidence of more adaptive computation. Other recent work measures step-level faithfulness through parameter-level unlearning or hint-injection paradigms~\citep{tutek2025fur,young2026lietome,young2026whymodelsknow}. Our contribution is more specific: a timing-based latent answer proxy cross-validated by multiple readouts, a five-way step-level taxonomy of mismatch patterns, architecture-matched Qwen2.5/DeepSeek-R1 comparisons across scale, and an empirical link between temporal faithfulness and CoT utility. A detailed axis-by-axis comparison against these and other closely related works is provided in~\Cref{app:differentiation}. Complementary work on normalized step-importance metrics such as NLDD targets cross-architecture comparability, whereas \bca{} is designed as a timing metric centered on when latent answer commitment becomes decodable relative to answer revelation~\citep{ye2026faithfulness_decay}.

\paragraph{Reasoning pipelines and training effects.}
Reasoning-oriented training pipelines, including process supervision, reinforcement learning, and distillation, increasingly optimize models to produce extended reasoning traces~\citep{uesato2022solving,cobbe2021training,zelikman2022star,havrilla2024teaching,deepseek2025r1,openai2024o1,hao2024coconut}. Whether these pipelines improve faithfulness rather than only usefulness remains unclear: longer or more polished traces may still fail to report the computation that determined the answer, and concurrent work has raised concerns about strategically uninformative or steganographic reasoning traces~\citep{roger2023preventing,chen2025reasoning}. Conversely, targeted interventions such as counterfactual simulation training can improve faithfulness in controlled settings, suggesting that the effect of training is real but objective-dependent~\citep{hase2026cst}. Our matched-family comparison targets this question directly by comparing Qwen2.5 models against their DeepSeek-R1-distilled counterparts at 7B, 14B, and 32B, which lets us study how the pipeline reshapes timing alignment and failure composition while minimizing backbone confounds.

\section{Method}
\label{sec:method}

\dcc{} follows a chain of thought one step at a time. For each step, we estimate whether the model has already committed to the final answer, compare that latent state with what the written trace has revealed so far, summarize the resulting mismatches with a taxonomy, and then use matched model pairs to study how reasoning training changes the alignment pattern.

\subsection{Step-Level Belief Extraction}

Given a model's CoT response to a question, we first segment the trace into reasoning steps $s_1, \ldots, s_T$ using a rule-based parser that splits on newlines, transition markers, and sentence boundaries (parsing F1 = 0.87; validation in \Cref{app:parsing}).

At each step boundary, we extract a proxy for the model's internal commitment to the final answer using the logit lens~\citep{nostalgebraist2020logitlens,chen2025logitlens4llms}. Concretely, we project each layer's hidden state through the unembedding matrix and layer normalization to obtain a distribution over the vocabulary. We use the probability assigned to the ground-truth answer token, $p_{\mathrm{ans}}^{(\ell)}(t)$, as our primary measure of internal commitment at step $t$ and layer $\ell$. This operationalization captures the dominant signal but may not reflect all aspects of the model's internal state; we discuss this limitation in \Cref{sec:limitations}.

We validate the logit lens through concordance with Patchscopes~\citep{ghandeharioun2024patchscopes} and tuned lens probes~\citep{alain2017understanding,belinkov2022probing}. Patchscopes agrees on 75.6\% of mid-depth steps and 74.7\% of final-layer steps, with near-zero logit-lens-only disagreement (0.0\% mid, 1.5\% final); tuned-lens binary classifications agree with the logit lens at $\geq$99.2\% across four held-out model-benchmark configurations (\Cref{app:logitlens}). Together these checks indicate that the logit lens behaves as a conservative answer-commitment readout rather than a trigger-happy artifact.

\subsection{Divergence Quantification}

We quantify timing alignment using two complementary metrics. \textbf{Belief-CoT Agreement} (\bca{}) measures the fraction of steps at which the thresholded answer-commitment proxy ($p_{\mathrm{ans}} > \tau$, $\tau = 0.3$) matches a binary CoT label indicating whether the trace has already explicitly reached the final answer. We compute that label automatically by checking, at each step, whether the ground-truth answer string or a normalized variant has appeared in the cumulative CoT text. A \bca{} of 1.0 indicates perfect timing alignment. Because the CoT label is binary, \bca{} should be read as a timing metric over answer commitment rather than a full semantic comparison; results are stable across $\tau \in [0.1, 0.5]$ (\Cref{app:threshold}). \textbf{Convergence Timing Gap} (\ctg{}) measures how many steps before the CoT arrives at the final answer the latent belief has already committed. A positive \ctg{} indicates premature internal convergence.

\subsection{Mismatch Taxonomy}
\label{sec:method-taxonomy}

Not all mismatches are alike. Based on temporal patterns in the answer-commitment trajectory, we classify detected mismatches into five heuristic categories. \textbf{Premature Convergence} (PC) occurs when the answer-commitment proxy locks onto the answer at least two steps before the trace states it. \textbf{Contradictory States} (CT) denote high-confidence disagreement between the proxy and the answer-arrival label. \textbf{Hidden Reasoning} (HR) captures large adjacent-step shifts in the proxy that are not mirrored by a change in explicit answer state. \textbf{Confabulated Steps} (CS) are mismatch steps where the proxy changes negligibly, suggesting that the trace continues after answer commitment has largely stabilized. We use ``confabulated'' as a descriptive label for this temporal pattern rather than as a strong claim about whether these steps serve a computational role that our proxy does not capture. \textbf{Silent Error Correction} (SEC) denotes runs of disagreement followed by re-alignment, consistent with an unspoken internal correction. See formal definitions and operational criteria in \Cref{app:taxonomy-def}.

These categories are not mutually exclusive: a single step or instance may satisfy multiple criteria (we analyze multi-label co-occurrence in \Cref{app:multilabel}). In the main summaries we count all detector firings, and when an instance-level summary is needed we report the dominant type by severity.

\subsection{Architecture-Controlled Pipeline Comparison}

A central challenge in studying how training affects faithfulness is confounding: different models usually differ in architecture, data, and training procedure at the same time. To isolate the effect of one concrete pipeline within a single backbone family, we leverage the DeepSeek-R1 distillation models, which use Qwen2.5 as their base architecture. This gives us matched pairs at three scales: Qwen2.5-7B vs.\ DS-R1-7B, Qwen2.5-14B vs.\ DS-R1-14B, and Qwen2.5-32B vs.\ DS-R1-32B. Within each pair, differences in \bca{} are therefore more plausibly linked to the distillation pipeline than to backbone changes, though the pipeline itself still bundles RL, SFT, and knowledge distillation. Our conclusions about pipeline effects are specific to the Qwen2.5/R1 setting; generalization to other backbone families remains open.

\subsection{Causal Validation via Direction Ablation}
\label{sec:causal-method}

The logit lens reveals correlational patterns in latent belief formation, but correlation does not establish that detected beliefs causally influence the model's output. To test causality, we perform \emph{direction ablation}~\citep{zou2023representation,goldowskydill2023localizing,geiger2024causal,vig2020investigating}: at a target layer $\ell$, we extract the answer-token direction $\hat{\mathbf{d}}_{a^*}$ from the unembedding matrix and project it out of the hidden state with strength $\alpha$:
\begin{equation}
    \tilde{\mathbf{h}}^{(\ell)} = \mathbf{h}^{(\ell)} - \alpha \cdot (\mathbf{h}^{(\ell)} \cdot \hat{\mathbf{d}}_{a^*})\,\hat{\mathbf{d}}_{a^*}
    \label{eq:ablation}
\end{equation}
We sweep $\alpha \in \{1, 3, 5, 10\}$ and report the strength that produces the largest change in answer probability, following the practice of selecting the most informative intervention point. If the belief component is causally relevant, removing it should change the model's output. We measure the \emph{correctness change rate}: the fraction of instances where the model's answer switches between correct and incorrect after ablation. We test four layers spanning 50--97\% of model depth on $n = 60$ instances per 7B configuration and $n = 20$ per 32B configuration, covering both the matched 7B pair and the matched 32B pair. Full details in \Cref{app:patching}.

\section{Experimental Setup}
\label{sec:experiments}

\subsection{Models}

We evaluate nine open-source models. Four primary models span distinct families: Llama-3.1-8B-Instruct~\citep{dubey2024llama}, Qwen3.5-9B~\citep{yang2025qwen3}, Gemma-3-27B-IT~\citep{team2025gemma3}, and DeepSeek-R1-Distill-Qwen-32B~\citep{deepseek2025r1} (abbreviated DS-R1-32B). For architecture-controlled analysis, we add Qwen2.5-Instruct models at 7B, 14B, and 32B~\citep{yang2024qwen25} with matched DeepSeek-R1-Distill-Qwen models at 7B and 14B~\citep{deepseek2025r1} (abbreviated DS-R1-7B and DS-R1-14B), yielding three Qwen2.5-to-R1 comparisons within a shared backbone (\Cref{app:models}). Not every analysis uses all models; we state the active count where it differs.

\subsection{Benchmarks}

We use seven reasoning benchmarks chosen to span distinct forms of multi-step inference: MATH-500~\citep{hendrycks2021math} for mathematical reasoning, PrOntoQA~\citep{saparov2023testing} for symbolic logic, ProsQA for planning, MMLU-Pro~\citep{wang2024mmlupro} for knowledge-intensive reasoning, BBH-Logical Deduction and BBH-Tracking Shuffled Objects~\citep{suzgun2023bbh} for compositional reasoning, and GPQA-Diamond~\citep{rein2024gpqa} for difficult scientific question answering. Depending on the benchmark, evaluation uses between 60 and 300 instances. \Cref{app:models} lists the exact instance counts and dataset sources.

\subsection{Inference and Measurement Setup}

All experiments use greedy decoding. We segment generated CoT traces with the rule-based step parser described in \Cref{app:parsing}, which achieves an F1 score of 0.87 on manually annotated traces. For belief extraction we read hidden states at every even-numbered layer and threshold the logit-lens answer probability at $\tau = 0.3$ when computing \bca{}, a choice that remains stable across $\tau \in [0.1, 0.5]$ (see \Cref{app:threshold}). Confidence intervals are 95\% bootstrap intervals with Bonferroni correction, and \bca{} is stratified by answer correctness to avoid conflating alignment with raw task difficulty. The full evaluation required approximately 200 GPU-hours on four A100 GPUs.  Temperature sensitivity checks (see \Cref{app:temperature}) and inference-time intervention experiments (see \Cref{app:interventions}) are reported in the appendix.

\section{Results and Analysis}
\label{sec:results}

This section examines whether the answer a model appears to commit to internally is synchronized with the point at which that answer becomes explicit in the written chain-of-thought trace. We use \bca{} as a step-level timing measure: high values indicate that latent answer commitment and explicit answer arrival move together, while low values indicate that the two are separated in time. Across models and benchmarks, the main pattern is simple. This separation is common, it varies strongly by task, and in many cases it is not visible from the surface text alone.

\subsection{Alignment Is Low, and Benchmark Choice Matters Most}
\label{sec:results-detection}

\begin{figure}[t]
\centering
\includegraphics[width=\columnwidth]{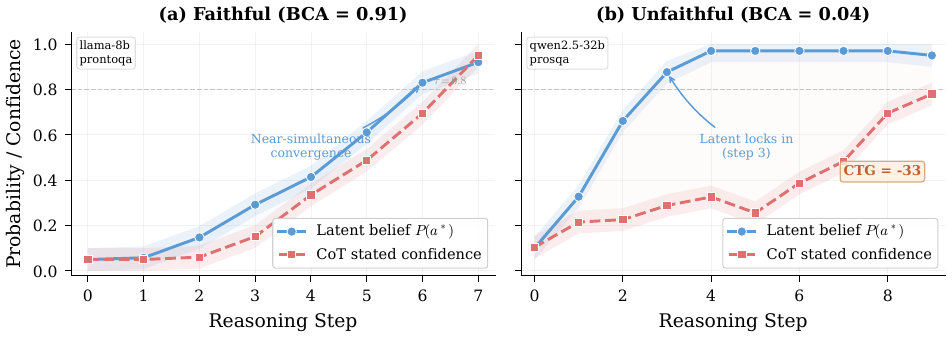}
\caption{Representative belief trajectories. \textbf{(a)} Faithful (Gemma 3 27B/GPQA-D, \bca{} $= 0.75$): latent belief and CoT converge simultaneously. \textbf{(b)} Unfaithful (Qwen 3.5 9B/GPQA-D, \bca{} $= 0.00$): latent belief locks by step~2 while CoT continues exploring, yielding a large convergence timing gap.}
\label{fig:trajectory-real}
\vspace{-2mm}
\end{figure}

\Cref{tab:main-results} reports \bca{} for each model-benchmark configuration. Averaged over all reported cells, \bca{} is 0.619. In practical terms, the latent commitment signal and the explicit answer in the CoT trace agree in timing on only about three-fifths of steps. The result does not look like an isolated failure of one model family. Among the four primary models, a one-way ANOVA does not detect a significant model effect ($F = 0.91$, $p = 0.45$), and Bayesian model comparison favors the null model ($\mathrm{BF}_{01} = 32.9$). This does not prove that architecture is irrelevant, but it does suggest that model identity alone cannot explain the observed timing gaps. A held-out Gemma 3 12B replication shows the same qualitative pattern of low and task-sensitive alignment, as reported in \Cref{app:gemma12b}.

\begin{table*}[t]
\centering
\small
\setlength{\tabcolsep}{4.6pt}
\begin{tabular}{@{}lccccccc|c@{}}
\toprule
\textbf{Model} & \textbf{MATH} & \textbf{PrOnto} & \textbf{ProsQA} & \textbf{MMLU-P} & \textbf{BBH-LD} & \textbf{BBH-TSO} & \textbf{GPQA-D} & \textbf{Mean} \\
\midrule
Llama 3.1 8B & \textbf{0.738} & \textbf{0.914} & 0.280 & 0.904 & 0.669 & 0.807 & 0.779 & 0.727 \\
Qwen 3.5 9B & 0.634 & 0.635 & 0.364 & 0.254 & 0.774 & 0.565 & 0.773 & 0.571 \\
Gemma 3 27B & 0.526 & 0.774 & 0.181 & 0.128 & 0.819 & 0.725 & 0.695 & 0.550 \\
DS-R1-32B & 0.614 & 0.529 & 0.329 & 0.231 & 0.538 & 0.890 & 0.769 & 0.557 \\
\midrule
\rowcolor{gray!10} Qwen2.5-7B & 0.579 & 0.807 & 0.307 & 0.895 & \textbf{0.871} & 0.804 & 0.808 & 0.724 \\
\rowcolor{gray!10} DS-R1-7B & 0.633 & 0.771 & 0.342 & \textbf{0.916} & 0.775 & 0.810 & 0.786 & 0.719 \\
\rowcolor{gray!10} Qwen2.5-14B & 0.581 & 0.496 & 0.236 & 0.193 & 0.704 & 0.508 & 0.789 & 0.501 \\
\rowcolor{gray!10} DS-R1-14B & 0.600 & 0.495 & \textbf{0.365} & 0.245 & 0.486 & \textbf{0.896} & \textbf{0.812} & 0.557 \\
\rowcolor{gray!10} Qwen2.5-32B & 0.627 & 0.678 & 0.044 & 0.897 & 0.827 & 0.783 & 0.783 & 0.663 \\
\bottomrule
\end{tabular}
\caption{Step-level answer-commitment alignment, measured by \bca{}. The largest differences are benchmark-level rather than model-level. Bold marks the highest value in each benchmark, and shaded rows denote the Qwen2.5 architecture-control pairs.}
\label{tab:main-results}
\vspace{-3mm}
\end{table*}

The benchmark effect is large. On PrOntoQA, where the reasoning structure is relatively explicit, Llama 3.1 8B reaches a \bca{} of 0.914. On other tasks, the same alignment measure can collapse. Qwen2.5-32B reaches only 0.044 on ProsQA, and Gemma 3 27B reaches 0.128 on MMLU-Pro. The representative trajectories in \Cref{fig:trajectory-real} show this contrast directly: in high-alignment settings, the answer-commitment curve and the written answer arrive together; in low-alignment settings, the model can appear committed well before the trace says so. A separate convergence-timing analysis points in the same direction. Lower \bca{} is associated with larger convergence timing gaps ($\rho = -0.48$, $p < 0.001$; \Cref{app:ctg}), which is consistent with latent commitment often preceding explicit answer arrival.

\subsection{Most Mismatches Occur After the Model Has Already Committed}
\label{sec:results-taxonomy}

We next ask what kind of behavior produces these timing gaps. We apply the five detectors defined in \Cref{app:taxonomy-def} to 168,302 mismatch events. The taxonomy should be read as a descriptive breakdown of overlapping failure patterns rather than as a set of mutually exclusive causes. With that caveat, one category dominates: Confabulated Steps (CS), where the answer-commitment proxy is already stable but the model continues producing CoT steps that induce little change in that proxy. CS accounts for 58.0\% of detector firings. The other categories are Hidden Reasoning (HR, 19.9\%), Contradictory States (CT, 15.5\%), Silent Error Correction (SEC, 5.2\%), and Premature Convergence (PC, 1.4\%).

\begin{figure}[t]
\centering
\includegraphics[width=0.85\columnwidth]{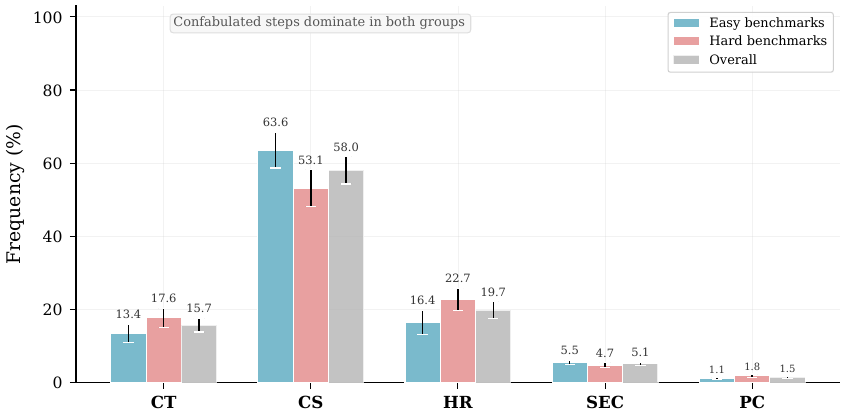}
\caption{Distribution of the five unfaithfulness types across easy and hard benchmarks (${\sim}168{,}000$ mismatch steps). Confabulated Steps dominate at all difficulty levels (58\% overall), indicating that most unfaithful reasoning is performative rather than erroneous.}
\label{fig:taxonomy-dist}
\vspace{-2mm}
\end{figure}

The dominance of CS is not driven by a single model. Across the tested models, per-model CS rates range from 37.4\% for DS-R1-32B to 75.4\% for Qwen-3.5-9B. Gemma 3 27B has an additional logit soft-capping artifact, discussed in \Cref{app:taxonomy-def}, but the pattern remains after accounting for it. Excluding Gemma 3 27B, CS still explains 56.5\% of the remaining 162{,}422 events. A held-out Gemma 3 12B replication under the same $\tau = 0.3$ threshold produces a non-degenerate taxonomy distribution and again shows a CS-dominant pattern (CS = 58.1\%; \Cref{app:gemma12b}).

This result changes the interpretation of the alignment gap. The typical failure is not that the written trace openly contradicts the internal answer signal. More often, the model has already settled on an answer and then continues writing plausible-looking intermediate steps. Several checks support this reading. In a vacuousness analysis over 314 instances containing CS steps, the committed answer never changes during a CS step, and 90.1\% of individual CS steps have near-zero belief change ($\Delta p_{\mathrm{ans}} < 0.001$; \Cref{app:cs-vacuousness}). A donor-corruption test gives a similar result from a different angle: replacing text at pure-CS steps leaves final-answer accuracy essentially unchanged ($\Delta = -0.002$, $p = 0.78$), while replacing matched neighbor or random steps is more damaging (\Cref{app:cs-corruption}). These controls make it less likely that CS is simply useful verification written after the answer.

Text-only inspection also misses much of this behavior. Three LLM judges, given only the CoT text, reach high agreement ($\mathrm{AC1} = 0.78$) but label 82\% of inspected cases as PC rather than CS (\Cref{app:judges}). Human annotators show moderate-to-high agreement on the five-way taxonomy ($\mathrm{AC1} = 0.71$; \Cref{app:human-annotation}). The gap between text-only labels and detector-based labels is important for oversight: many post-commitment steps still look like normal reasoning.

The categories also overlap substantially. In the 12{,}004-instance audit pool, 68.4\% of unfaithful steps satisfy more than one detector criterion (\Cref{app:multilabel}). CT and PC form a near-redundant pair (Pearson $r = 0.937$, 95\% CI $[+0.930, +0.943]$; \Cref{tab:detector-correlation}), and neither PC nor SEC appears as an isolated category in that pool. The taxonomy is still useful, but mainly as a way to describe correlated facets of trace unfaithfulness.

The cleanest causal evidence comes from paired truncation. Among the categories with non-empty pure subgroups, cutting the trace at a pure-CS firing step improves accuracy relative to cutting at a uniformly random step by $+0.078$ ($n = 2{,}088$, $p < 10^{-16}$; \Cref{app:truncation-v3}). If those CS steps were load-bearing reasoning steps, truncating at them should not help. HR behaves differently: pooled deltas are positive rather than negative, suggesting that many large belief jumps mark the completion of commitment rather than hidden computation that remains necessary afterward. CT trends in the opposite direction, but the pure-CT subgroup is too small to support a strong conclusion. The detector thresholds are stable under $\pm 50\%$ perturbations (\Cref{app:threshold-xval}), and 85\% of CS events occur without premature convergence (\Cref{app:cs-pc}).

\subsection{The Reasoning Pipeline Changes the Type of Mismatch}
\label{sec:results-pipeline}

We next compare DeepSeek-R1 distilled models with their Qwen2.5 architecture controls. Since each pair shares the same backbone size, this comparison gives a cleaner view of the reasoning pipeline than the cross-family results above. The aggregate effect on \bca{} is mixed rather than monotonic: alignment decreases at 32B (mean $\Delta_{\text{pipe}}=-0.106$), improves at 14B ($+0.056$), and is nearly unchanged at 7B ($-0.005$). The 7B-versus-32B contrast is not significant under a paired Wilcoxon test ($p=0.578$), and the 32B mean becomes much smaller once the MMLU-Pro outlier is removed ($\Delta_{\text{pipe}}=-0.012$ without the $-0.666$ MMLU-Pro cell; full per-benchmark deltas in \Cref{tab:pipeline}). The aggregate numbers should therefore not be read as a simple scaling law.

The taxonomy shows a clearer pipeline effect than \bca{} alone. Even when the overall alignment score changes only modestly, the distribution of mismatch types can change substantially. Chi-squared tests on the five-way taxonomy distribution show increasing divergence with scale, with Cram\'{e}r's $V$ rising from 0.058 at 7B to 0.155 at 14B and 0.289 at 32B (\Cref{app:taxonomy-shift}). At 7B, the taxonomy distribution is largely preserved, with CS moving from 53.0\% in Qwen2.5-7B to 57.9\% in DS-R1-7B. At 14B, the pipeline reduces the average number of mismatches per instance from 24.2 to 14.7, a 39\% reduction that matches the positive \bca{} effect at that scale.

The strongest shift appears at 32B. Confabulated steps fall from 57.6\% to 37.4\%, while contradictory states rise from 7.1\% to 28.9\%. This is not only a change in how often mismatch occurs. It changes what a monitor would need to detect. A monitor tuned for post-commitment continuation may miss high-confidence disagreements between the internal answer signal and the written trace. DS-R1-32B on ProsQA illustrates this point sharply: the model reaches 98.3\% accuracy, but its \bca{} is only 0.329, and none of the analyzed instances are classified as faithful.

\subsection{Belief Commitment Appears Late in the Forward Pass}
\label{sec:results-mech}

The layer-wise logit-lens analysis gives a compatible picture of the timing results. Across models, answer probability is close to chance in early layers, rises through the middle of the network, and often increases sharply near the final layers. The DeepSeek-R1 pipeline amplifies this late-layer pattern at 32B, while the corresponding gap at 7B is much smaller. This mirrors the behavioral results above: at larger scale, the pipeline appears to concentrate answer resolution into later layers that are less tightly synchronized with the point at which the answer appears in the CoT trace.

\begin{figure}[t]
\centering
\includegraphics[width=0.95\columnwidth]{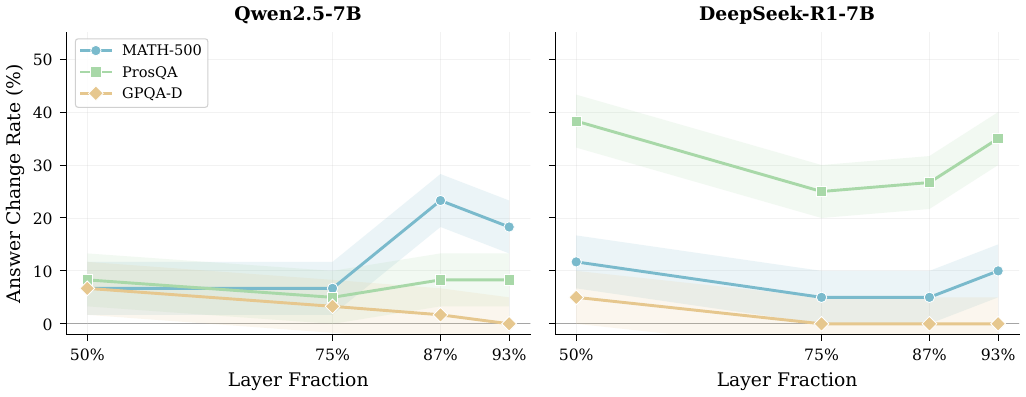}
\caption{Direction ablation at selected layer depths. Removing the answer-token direction changes correctness most strongly in late layers for several configurations, with the largest observed effect in DS-R1-32B on ProsQA.}
\label{fig:patching}
\vspace{-3mm}
\end{figure}

Direction ablation provides a stronger check that the detected answer-commitment signals are not merely probe artifacts. We project out the answer-token direction from hidden states at four layer depths, as described in \Cref{sec:causal-method}. At 7B, this changes correctness in up to 23\% of Qwen2.5-7B/MATH-500 cases and up to 38\% of DS-R1-7B/ProsQA cases. At 32B, DS-R1-32B on ProsQA reaches a 55\% correctness change, with a clear late-layer peak (\Cref{app:patching}). These results support the causal relevance of the answer-commitment direction, but they do not establish a universal late-layer rule across all tasks. Patchscopes concordance and tuned-lens agreement provide additional evidence that the logit lens is a conservative estimator of answer commitment rather than an idiosyncratic probe (\Cref{app:logitlens}).

\subsection{CoT Helps Most Where Timing Alignment Is Weak}
\label{sec:results-utility}

Finally, we ask whether low alignment is related to the practical value of CoT. For the five models with direct no-CoT baselines, we compute CoT utility as the accuracy gain from using CoT, giving 35 model-benchmark configurations. \bca{} has a moderate negative Pearson correlation with CoT utility ($r = -0.42$, $p = 0.012$; bootstrap 95\% CI $[-0.64, -0.13]$; \Cref{app:utility-scatter}). In other words, the settings where CoT helps most are often the settings where step-level timing alignment is weakest. All five within-model correlations are negative ($r \in [-0.52, -0.39]$), although each individual correlation is underpowered at $n = 7$.

The relationship is not uniform. The Spearman correlation is directionally consistent but not significant ($\rho = -0.25$, $p = 0.16$), and the leave-one-benchmark-out analysis identifies ProsQA as the main leverage point (\Cref{app:utility-sensitivity}). This sensitivity is informative rather than purely problematic. ProsQA is also the clearest graph-planning setting in the benchmark suite, where CoT is highly useful but step-level alignment is very low. The result should be read as a moderate association, not as a universal law. As a final check, \bca{} is nearly uncorrelated with an early-commitment rate derived from answer-probability trajectories ($\rho = 0.05$, $p = 0.73$, $n = 56$; \Cref{app:cross-method}), which suggests that \bca{} captures timing information not reducible to response-level early answering.

Taken together, the results show that CoT unfaithfulness is often a timing problem rather than a content-only problem. The model may commit to an answer before the written reasoning arrives there, and the remaining trace can still look coherent. This makes surface-level CoT inspection unreliable in precisely the settings where CoT is most useful.
\section{Discussion}
\label{sec:discussion}

The paired-truncation analysis turns the descriptive taxonomy into an intervention-based test (\Cref{app:truncation-v3}). Cutting at a confabulated step more often preserves or improves final-answer correctness than cutting at a matched control step from the same instance, which supports the interpretation of many such segments as post-commitment continuation rather than load-bearing reasoning. The corresponding hidden-reasoning result is more ambiguous: pooled deltas are positive, suggesting that many large belief jumps mark commitment completion rather than latent computation that still needs to be verbalized. CT remains directionally load-bearing but underpowered, while SEC and PC are structurally non-isolable as pure subgroups. We therefore use the five-way taxonomy as a descriptive decomposition of correlated mismatch patterns, not as five independent causal mechanisms.

These findings matter most for oversight. Process reward models and step-level monitors may reward fluent continuation after commitment rather than the computation that determines the answer~\citep{lightman2023verify}. The matched-family comparison further shows that training pipelines can change which failures dominate even when aggregate alignment moves little, consistent with broader evidence that CoT monitorability is sensitive to development choices~\citep{baker2025cot_monitorability}. The negative association between alignment and CoT utility points to a narrower tension: the settings in which CoT helps most may also be the settings in which the trace is least informative about answer formation, although in our data this pattern is driven most strongly by ProsQA-like graph-planning tasks. We do not conclude that long CoT is generally empty or useless; some harder tasks still show genuine mid-trace belief shifts~\citep{boppana2026reasoning_theater}. The narrower conclusion is that CoT should not be assumed to report, step by step, when the answer was formed. If CoT is to function as an oversight interface, future work will need methods that distinguish load-bearing reasoning from post-commitment narration and training procedures that make that distinction easier to observe.

\section{Conclusion}

We studied whether chain-of-thought traces faithfully report when a model forms its answer. Across nine open-weight models and seven reasoning benchmarks, latent answer commitment and explicit answer arrival are often misaligned, with post-commitment continuation as the dominant mismatch pattern. Matched Qwen2.5/DeepSeek-R1 comparisons show that reasoning pipelines can change the type of unfaithfulness even when aggregate alignment changes only modestly. The main lesson is not that CoT is useless, but that it is an imperfect oversight channel: a trace can improve accuracy while failing to reveal when the answer was actually formed. Future process monitors should therefore distinguish load-bearing reasoning from plausible post-commitment narration.

\section{Limitations}
\label{sec:limitations}

Our $p_{\mathrm{ans}}$ proxy captures answer-commitment \emph{timing} rather than full semantic equivalence, and may underestimate beliefs encoded nonlinearly or across multiple tokens (mitigated via Patchscopes and tuned-lens concordance; \Cref{app:logitlens}). The causal analysis reports the strongest effect across intervention strengths, possibly overstating single-setting sizes. The taxonomy is overlapping by design (\Cref{app:multilabel}), the scale-dependent reversal holds only within the Qwen2.5/DeepSeek-R1 family, and all models are open-weight (7B--32B).

\bibliography{ref}
\bibliographystyle{plainnat}

\appendix
\section{Model and Benchmark Specifications}
\label{app:models}

\Cref{tab:models-full} provides the full architectural specifications for all models used in this study.

\begin{table}[h]
\centering
\small
\begin{tabular}{@{}lcccc@{}}
\toprule
\textbf{Model} & \textbf{Params} & \textbf{Layers} & \textbf{$d_{\text{model}}$} & \textbf{Backbone} \\
\midrule
Llama 3.1 8B-Inst. & 8.0B & 32 & 4096 & Llama 3.1 \\
Qwen 3.5 9B & 9.0B & 40 & 3584 & Qwen 3.5 \\
Gemma 3 27B-Inst. & 27.2B & 62 & 5376 & Gemma 3 \\
DS-R1-Qwen-32B & 32.5B & 64 & 5120 & Qwen2.5 \\
\midrule
\rowcolor{gray!10} Qwen2.5-7B-Inst. & 7.6B & 28 & 3584 & Qwen2.5 \\
\rowcolor{gray!10} DS-R1-Qwen-7B & 7.6B & 28 & 3584 & Qwen2.5 \\
\rowcolor{gray!10} Qwen2.5-14B-Inst. & 14.7B & 48 & 5120 & Qwen2.5 \\
\rowcolor{gray!10} DS-R1-Qwen-14B & 14.7B & 48 & 5120 & Qwen2.5 \\
\rowcolor{gray!10} Qwen2.5-32B-Inst. & 32.5B & 64 & 5120 & Qwen2.5 \\
\bottomrule
\end{tabular}
\caption{Model specifications. Architecture-controlled baselines are shaded.}
\label{tab:models-full}
\end{table}

\Cref{tab:benchmarks-full} provides benchmark specifications.

\begin{table}[h]
\centering
\small
\begin{tabular}{@{}lcccl@{}}
\toprule
\textbf{Benchmark} & $N_{\text{eval}}$ & $N_{\text{probe}}$ & \textbf{Steps} & \textbf{Type} \\
\midrule
MATH-500 & 300 & 130 & 4--8 & Mathematical \\
PrOntoQA & 300 & 130 & 5 & Logical \\
ProsQA & 60--88$^\dagger$ & 28 & 3--6 & Graph planning \\
MMLU-Pro & 300 & 130 & 2--5 & Knowledge \\
BBH-LD & 200 & 50 & 3--5 & Compositional \\
BBH-TSO & 200 & 50 & 3--7 & State tracking \\
GPQA-D & 130 & 68 & 3--6 & Science \\
\bottomrule
\end{tabular}
\caption{Benchmark specifications. $N_{\text{eval}}$ is the evaluation sample size; $N_{\text{probe}}$ is the probe training split size. $^\dagger$ProsQA uses $N_{\text{eval}}=88$ for Llama 3.1 8B, Qwen 3.5 9B, Qwen2.5-7B, and DS-R1-7B; $N_{\text{eval}}=60$ for the remaining five models.}
\label{tab:benchmarks-full}
\end{table}

\section{Pipeline Comparison Visualization}
\label{app:pipeline-fig}

\begin{figure*}[h]
\centering
\includegraphics[width=\textwidth]{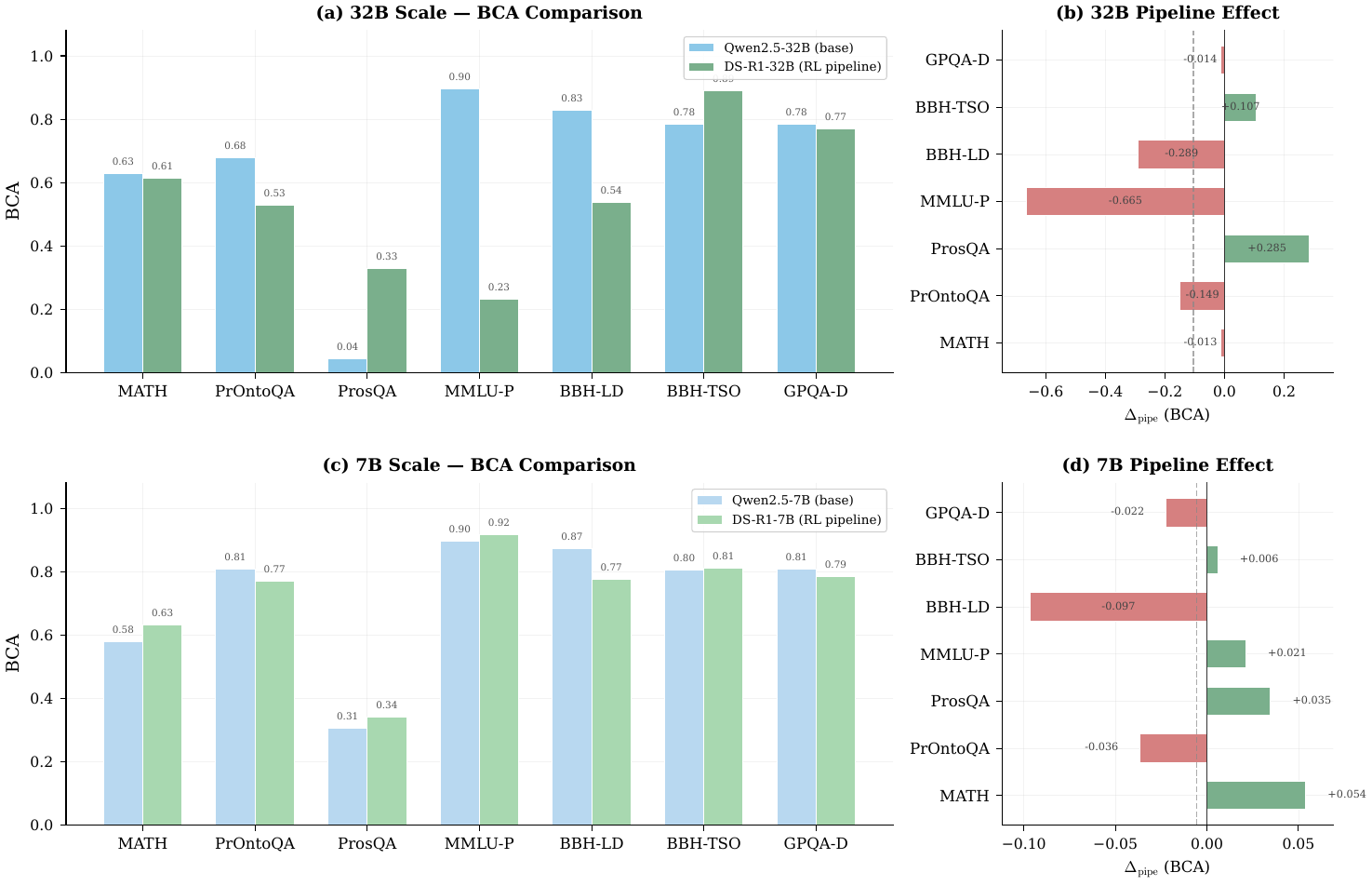}
\caption{Architecture-controlled pipeline comparison. Left panels show \bca{} for Qwen2.5 (base) and DS-R1 (pipeline-trained) at 32B and 7B; right panels show the per-benchmark effect $\Delta_{\text{pipe}}$. The main pattern is scale-dependent: the pipeline reduces alignment on most 32B benchmarks, while the net 7B effect is close to zero.}
\label{fig:pipeline}
\end{figure*}

\section{Mismatch Taxonomy: Definitions and Structural Analysis}

\subsection{Definitions}
\label{app:taxonomy-def}

\Cref{tab:taxonomy-def} provides the operational definitions and detector criteria used in the current implementation. All thresholds are calibrated on 50 held-out instances and validated through $\pm 50\%$ perturbation analysis.

\begin{table*}[h]
\centering
\small
\setlength{\tabcolsep}{3.8pt}
\begin{tabular}{@{}llp{5.7cm}p{3.8cm}@{}}
\toprule
\textbf{Type} & \textbf{Abbr.} & \textbf{Definition} & \textbf{Criterion} \\
\midrule
Premature Convergence & PC & Answer-commitment proxy locks onto the answer $\geq 2$ steps before the trace states it & $\ctg \geq 2$ \\
Contradictory States & CT & High-confidence disagreement between the proxy and the answer-arrival label & $\bca = 0 \;\wedge\; \text{probe confidence} \geq 0.5$ \\
Hidden Reasoning & HR & Large adjacent-step shift in answer-commitment proxy with no change in explicit answer state & Main classifier: $|\Delta p_{\text{ans}}| > 0.15$ on aligned steps \\
Confabulated Steps & CS & Mismatch step where answer probability changes negligibly across adjacent steps & $\Delta p_{\text{ans}} < 0.02 \;\wedge\; \bca = 0$ \\
Silent Error Correction & SEC & Run of disagreement followed by re-alignment without explicit revision in the trace & $\geq 2$ disagreement steps before re-alignment \\
\bottomrule
\end{tabular}
\caption{Mismatch taxonomy: formal definitions and operational criteria.}
\label{tab:taxonomy-def}
\end{table*}

\paragraph{Gemma 3 27B taxonomy note.} Gemma~3 27B uses logit soft-capping, which compresses output probabilities and causes the HR, CT, and SEC detectors to fall below their activation thresholds on all benchmarks. As a result, all 5{,}880 Gemma 3 27B mismatch events are classified as CS. This is a \emph{model-specific} artifact of the 27B configuration, not a property of the Gemma family or of the \bca{} framework: our held-out Gemma 3 12B replication (\Cref{app:gemma12b}) produces a full, non-degenerate taxonomy distribution under the same $\tau = 0.3$ threshold, with CS at 58.1\% in line with the overall 58.0\% figure. The aggregate conclusion is unaffected: excluding Gemma 3 27B, CS still accounts for 56.5\% of the remaining 162{,}422 events and remains the dominant category in all other eight models. Normalized alternatives such as NLDD~\citep{ye2026faithfulness_decay} offer a complementary lens focused on cross-architecture comparability.

\subsection{Multi-Label Analysis}
\label{app:multilabel}

We perform a focused multi-label audit on the 28 configurations used in the original four-model analysis, re-evaluating 9,613 unfaithful steps with all five detectors applied independently. Because activations were not cached for all 28 configurations, this appendix analysis reconstructs answer-commitment trajectories from CoT-only heuristics and should be interpreted as a robustness audit rather than the main estimate. In this audit, 68.4\% of unfaithful steps satisfy multiple taxonomy criteria simultaneously. The most frequent co-occurrence is PC + SEC (5,897 steps), followed by CT + PC (3,310) and CT + SEC (3,182). This confirms that the taxonomy should be interpreted as a set of overlapping failure dimensions rather than a strictly exclusive partition: disagreement between nearby categories such as PC and CS often reflects genuine phenomenological overlap rather than annotation noise.

\begin{figure}[h]
\centering
\includegraphics[width=0.95\textwidth]{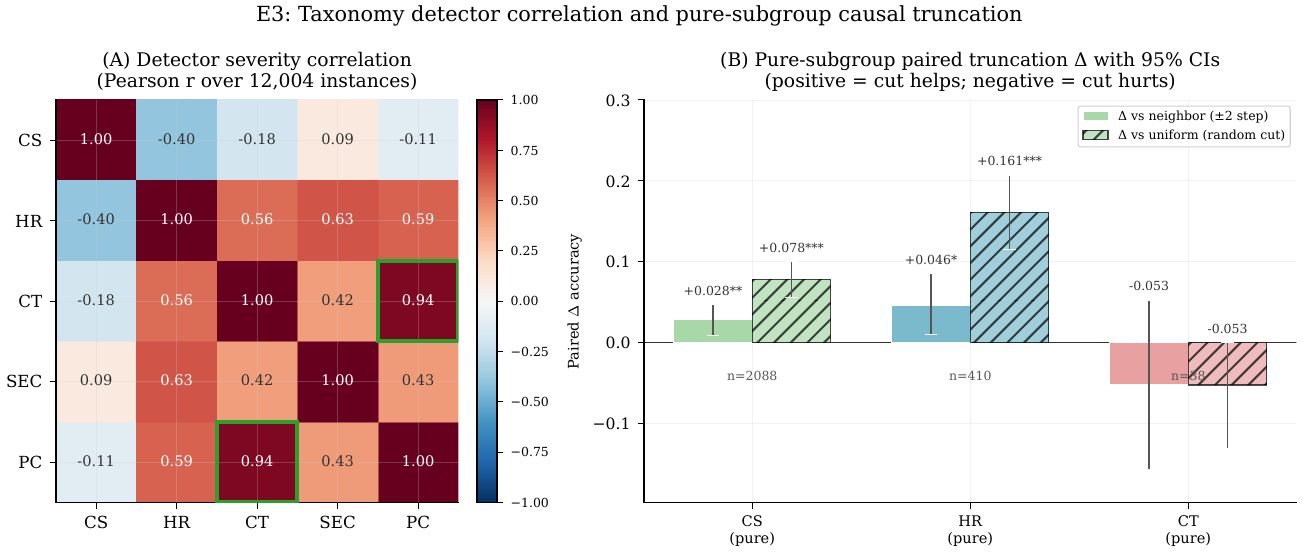}
\caption{\textbf{(A)} Pearson correlation matrix over 12{,}004 instance-level peak severities. The CT$\leftrightarrow$PC pair is highlighted: $r=0.937$, 95\% CI $[+0.930, +0.943]$. \textbf{(B)} Pure-subgroup paired $\Delta$ (treatment minus control) for the three active categories with 95\% bootstrap CIs. Hatched bars show $\Delta_{\text{vs U}}$ (uniform random cut); solid bars show $\Delta_{\text{vs N}}$ (matched neighbor $\delta=\pm 2$ step). CS and HR both show significant positive $\Delta$, the opposite of the load-bearing prediction for HR. CT is directionally negative but underpowered ($n=38$).}
\label{fig:e3-pure-truncation}
\end{figure}

\paragraph{Detector severity correlation structure.}
To quantify how overlapping the five detectors are, we re-run the full detector pipeline on 12{,}004 evaluation instances from 56 model-benchmark configurations (8 open-weight models $\times$ 7 benchmarks) and compute the Pearson correlation matrix over per-instance peak severities (\Cref{fig:e3-pure-truncation}~A):

\begin{table}[h]
\centering
\small
\begin{tabular}{lccccc}
\toprule
 & CS & HR & CT & SEC & PC \\
\midrule
CS  & $+1.000$ & $-0.395$ & $-0.180$ & $+0.095$ & $-0.108$ \\
HR  & $-0.395$ & $+1.000$ & $+0.561$ & $+0.628$ & $+0.586$ \\
CT  & $-0.180$ & $+0.561$ & $+1.000$ & $+0.421$ & $\mathbf{+0.937}$ \\
SEC & $+0.095$ & $+0.628$ & $+0.421$ & $+1.000$ & $+0.428$ \\
PC  & $-0.108$ & $+0.586$ & $\mathbf{+0.937}$ & $+0.428$ & $+1.000$ \\
\bottomrule
\end{tabular}
\caption{Pearson correlation matrix over 12{,}004 instance-level peak severities. The {HR, CT, SEC, PC} cluster is internally correlated at mean $r = +0.594$ (range $[+0.421, +0.937]$), while CS is semi-independent (mean $r = -0.147$ with the other four). CT and PC are nearly redundant: $r = 0.937$ with 95\% bootstrap CI $[+0.930, +0.943]$.}
\label{tab:detector-correlation}
\end{table}

Two observations follow. \emph{First}, the five-way taxonomy is most naturally described as \emph{two} dimensions rather than five independent categories: a \emph{commitment-state cluster} $\{$HR, CT, SEC, PC$\}$ whose severities move together, and a \emph{quiescent-tail signal} (CS) that varies semi-independently. \emph{Second}, CT and PC are nearly perfectly correlated ($r = 0.937$), and any future refinement of the taxonomy should consider merging them or treating them as a single two-step descriptor.

\paragraph{Structural non-isolability of PC and SEC.}
A consequence of the correlation structure is that the PC and SEC detectors \emph{never fire in isolation} in our evaluation pool. At a severity threshold of $0.30$, 2{,}123 instances trigger PC and 2{,}320 instances trigger SEC across the 12{,}004-instance pool, yet \emph{zero} of those instances have only PC or only SEC above threshold—100\% of PC firings co-occur with at least one of $\{$CS, HR, CT, SEC$\}$, and likewise for SEC. This is not a sample-size artifact: even at $7\times$ the original E2c pool size, the number of pure-PC and pure-SEC instances remains exactly zero. Structurally:
\begin{itemize}[leftmargin=*,nosep]
\item PC measures the gap between latent commitment and explicit answer arrival. Any PC-firing instance must contain post-commitment steps, which automatically register as CS (low $|\Delta p_{\mathrm{ans}}|$) or HR (a commitment jump).
\item SEC measures a disagreement run followed by recovery. The disagreement window by construction triggers CT-like steps during its body.
\end{itemize}
We therefore report the five-way taxonomy as an overlapping descriptive decomposition and explicitly do \emph{not} claim that PC and SEC are individually causally testable at the step level; our causal-truncation analysis (\Cref{app:truncation-v3}) is run only on the three categories that have non-empty pure subgroups (CS, HR, CT).

In the 12{,}004-instance pool, 73.1\% of instances have exactly one detector firing above threshold, and only 8.7\% have all five firing simultaneously. The earlier 68.4\% multi-label figure was computed on the top-severity stratum of the original 28 configurations and overstates the marginal co-firing rate in the full evaluation population.

\subsection{Confabulation Independence from Premature Convergence}
\label{app:cs-pc}

A natural concern is whether Confabulated Steps (CS) are merely a downstream consequence of Premature Convergence (PC): if the model commits early, subsequent steps will mechanically show small $\Delta p_{\mathrm{ans}}$. To test this, we partition CS detections by whether the enclosing instance also triggered the PC detector ($\ctg \geq 2$). Across all 70 model-benchmark configurations (103,218 total CS events), an estimated 85.4\% of CS events occur in instances that do \emph{not} trigger PC. Per-model rates range from 63.5\% (DS-R1-32B, which has the highest PC prevalence at 44.1\% of instances) to 100\% (Gemma 3 27B, which has zero PC instances). Per-benchmark rates range from 53.4\% (GPQA-Diamond) to 99.4\% (MMLU-Pro). This indicates that confabulation is not reducible to premature convergence: the majority of CS events arise in instances where the model has \emph{not} committed to its answer unusually early by the PC criterion, yet individual steps still show negligible change in the answer-commitment proxy.

\section{Threshold Sensitivity and Cross-Family Validation}

\subsection{Threshold Sensitivity Analysis}
\label{app:threshold}

We validate that \bca{} is robust to the choice of belief threshold $\tau$ by sweeping across $\tau \in \{0.1, 0.15, 0.2, 0.25, 0.3, 0.35, 0.4, 0.45, 0.5\}$ on 56 model-benchmark configurations (8 models $\times$ 7 benchmarks; Gemma 3 27B is excluded because its logit soft-capping collapses all detectors into CS at every threshold---a model-specific artifact of the 27B configuration discussed in \Cref{app:taxonomy-def}). Mean \bca{} varies by less than 0.003 across the entire range (0.5555--0.5578). All qualitative conclusions hold at every threshold value: the model ranking is perfectly preserved, the benchmark difficulty ordering is stable, and the per-model and per-benchmark curves are nearly flat. The held-out Gemma 3 12B replication (\Cref{app:gemma12b}) further confirms that $\tau = 0.3$ generalizes within the Gemma family, reproducing the main CS-dominant pattern on a model that was not part of the calibration set. This demonstrates that our findings are not an artifact of the threshold choice.


\subsection{Cross-Family Threshold Validation}
\label{app:threshold-xval}

To validate that taxonomy thresholds generalize across model families rather than overfitting to the calibration set, we partition models into two families: Family~A (Llama 3.1 8B, Qwen 3.5 9B; non-Qwen2.5 backbone) and Family~B (Qwen2.5-\{7B, 14B, 32B\}, DeepSeek-R1-\{7B, 14B, 32B\}; Qwen2.5 backbone). We compute the taxonomy distribution for each family and measure the maximum category proportion shift at the default thresholds and under $\pm 25\%$ and $\pm 50\%$ threshold perturbation. At default thresholds, the maximum shift is 3.4\% (PC category), well below the 10\% generalization criterion. Under the most extreme perturbation ($\pm 50\%$), the worst-case shift remains only 3.48\%. This confirms that the taxonomy thresholds calibrated on a small held-out set generalize reliably across model architectures.

\subsection{Held-Out Gemma 3 12B Replication}
\label{app:gemma12b}

As an additional robustness check, we run the full E1 pipeline on a held-out Gemma 3 12B model that was not included in the main nine-model analysis. This replication reproduces the main qualitative pattern: alignment remains low and highly task-dependent (mean \bca{} = 0.570 across seven benchmarks; range 0.101 on MMLU-Pro to 0.926 on ProsQA), and confabulated steps again account for 58.1\% of all mismatch events, statistically indistinguishable from the overall 58.0\% figure. Critically, Gemma 3 12B produces a full, non-degenerate taxonomy distribution (HR, CT, and SEC detectors all fire) under the same $\tau = 0.3$ threshold used throughout the paper. This shows that the detector collapse observed in Gemma 3 27B (\Cref{app:taxonomy-def}) is specific to the 27B configuration rather than a family-wide property of Gemma, and that the \bca{} framework and its absolute-threshold instantiation generalize on a held-out out-of-family model.

\section{LLM Judge Validation}
\label{app:judges}

We conduct a multi-annotator validation study using three independent LLM judges from different model families (Llama-3.1-70B-Instruct~\citep{dubey2024llama}, Qwen3.5-27B~\citep{yang2025qwen3}, Qwen3-30B-A3B~\citep{yang2025qwen3}) on 300 stratified validation instances. Each judge receives the CoT text together with summary signals derived from the current pipeline and independently assigns a taxonomy label. Key agreement statistics are as follows:

\begin{itemize}[nosep]
    \item \textbf{Pairwise exact agreement}: 77--82\% across all three judge pairs
    \item \textbf{All-3 exact agreement}: 71.0\%
    \item \textbf{Gwet's $\mathrm{AC1}$ (nominal)}: 0.7804
    \item \textbf{Fleiss' $\kappa$}: 0.362 on five-way labels
    \item \textbf{Majority-vote agreement with seeded algorithmic label}: 24\%
\end{itemize}

Human validation results are reported separately in \Cref{app:human-annotation}.

The gap between Gwet's $\mathrm{AC1}$ and Fleiss' $\kappa$ is driven largely by the extreme imbalance of the judged label distribution. Across all judge labels, 82.0\% are PC, while CS, HR, and CT each account for only about 5.8--6.0\%, and SEC accounts for just 0.3\%. Under such prevalence skew, chance-corrected coefficients such as Fleiss' $\kappa$ and Krippendorff's nominal $\alpha$ are known to be conservative, whereas Gwet's $\mathrm{AC1}$ is substantially more stable~\citep{gwet2008computing}. For this reason, we report Gwet's $\mathrm{AC1}$ as our primary prevalence-robust summary of nominal inter-judge agreement, and treat Fleiss' $\kappa$ as a complementary statistic that highlights the severity of the imbalance rather than a contradiction of the underlying pairwise and all-judge agreement rates.

\paragraph{The text--activation gap as an additional finding.}
The most informative result of the judge study is the systematic \emph{disagreement} between text-based judges and the activation-based taxonomy. LLM judges who inspect only the CoT text and summary signals achieve high inter-judge consistency ($\mathrm{AC1} = 0.78$), yet they systematically disagree with the activation-based labels (24\% majority-vote agreement), overwhelmingly classifying instances as PC (82\%) while the activation analysis reveals CS dominates at the step level (58\%). This gap itself provides evidence for our central thesis: confabulated steps are \emph{performative}---they look like legitimate deliberation to text-based inspectors precisely because the surface reasoning is fluent and coherent, making this failure mode harder to detect without access to internal states. In other words, the low text--activation agreement is not a failure of the taxonomy but a demonstration of the oversight limitation the taxonomy is designed to measure.

Three structural factors further explain the distribution mismatch. First, judges assign a single dominant label per \emph{instance}, whereas the main taxonomy counts \emph{step-level} detector firings. An instance judged as PC (because it committed early) typically contains many subsequent CS steps (each showing negligible $\Delta p_{\mathrm{ans}}$). This granularity difference explains why step-level CS dominates (58\%) even though instance-level PC is frequent in judge labels: a single PC instance can generate dozens of CS step-firings. Second, the taxonomy is intentionally overlapping: 68.4\% of unfaithful steps satisfy multiple criteria simultaneously in the multi-label audit (\Cref{app:multilabel}), so disagreement between adjacent categories such as PC and CS often reflects genuine phenomenological overlap rather than arbitrary noise. Third, the validation pool is not human-gold-labeled; it is stratified from algorithmically seeded validation instances.

\section{Step Parsing Validation}
\label{app:parsing}

The CoT decomposer identifies step boundaries using transition markers (``therefore,'' ``so,'' ``thus''), numbering patterns, and sentence-final punctuation. Validation on 350 CoT traces against human annotations yields: step count agreement within $\pm 1$ for 91.1\% of traces, boundary $F_1 = 0.87$ (range 0.81--0.95), and \bca{} sensitivity below 0.02 when substituting human boundaries. Taxonomy classification changes for fewer than 4\% of steps under human boundaries.

\section{Convergence Timing Analysis}
\label{app:ctg}

\ctg{} distributions reveal distinct patterns across task types. Reasoning tasks (MATH, BBH) show predominantly positive \ctg{} (premature convergence), with values reaching 24.1 steps on BBH-TSO for Qwen 3.5 9B. Knowledge-retrieval tasks (MMLU-Pro) show negative \ctg{}, consistent with models retrieving answers before generating supporting reasoning. The correlation $\rho(\bca{}, \ctg{}) = -0.48$ ($p < 0.001$) confirms that larger convergence timing gaps systematically predict lower alignment.

\section{Mechanistic Intervention Analysis}

\subsection{Intervention Analysis}
\label{app:interventions}

We test two lightweight inference-time interventions on $n = 40$ instances per benchmark.

\textbf{Self-Verification (SV):} Appending ``Wait, let me verify each step\ldots'' after the initial CoT and generating a correction pass~\citep{madaan2023selfrefine,huang2024self}. SV produces significant gains on weakly committed tasks with verifiable structure: Llama 3.1 8B on GPQA-D improves from 0\% to 40\% ($p < 0.001$), Qwen 3.5 9B on GPQA-D from 10\% to 62.5\% ($p < 0.001$), and Qwen 3.5 9B on BBH-LD from 27.5\% to 77.5\% ($p < 0.001$). However, SV can harm performance on tasks where the model has already committed strongly: PrOntoQA for Llama drops from 87.5\% to 65.0\% ($p = 0.042$).

\textbf{Majority-Vote (MV):} Generating $N = 3$ responses at $T = 0.7$ and selecting the majority answer~\citep{wang2023selfconsistency}. MV is generally ineffective or harmful, with DS-R1-32B on BBH-TSO showing catastrophic degradation from 85\% to 27.5\% ($-57.5$ pp). This suggests that the pipeline-trained model's reasoning is fragile to stochastic perturbation.

Aggregate results: SV mean improvements are Qwen 3.5 9B $+20.0$ pp, DS-R1-32B $+9.6$ pp, Llama 3.1 8B $+3.2$ pp, Gemma 3 27B $+0.7$ pp. MV mean effects are uniformly near zero or negative.

\subsection{Mechanistic Analysis Details}
\label{app:mechanistic}

Layer-wise analysis probes hidden states at every even-numbered layer. For each configuration, we compute $p_{\text{ans}}$ at each probed layer and measure the ratio between the final layer's probability and the first layer's probability (amplification ratio). Results across all configurations:

Llama 3.1 8B: amplification ranges from $8\times$ (MMLU-Pro) to $125\times$ (ProsQA), with $>80\%$ of buildup concentrated in the final 13\% of layers for 12/14 configurations.

Qwen 3.5 9B: amplification ranges from $516\times$ (MMLU-Pro) to $2{,}466\times$ (MATH-500). Mann-Whitney $U$ test confirms that final-phase probability is significantly higher than plateau-phase probability on 17/21 configurations ($p < 0.05$).

DS-R1-32B: amplification reaches $32{,}330\times$ (PrOntoQA) and $65{,}600\times$ (MATH-500) on reasoning tasks, but remains $<1\times$ on MMLU-Pro. The task-selectivity suggests the pipeline amplifies late-layer commitment specifically for tasks where step-level verification was part of training.

The correlation between amplification and \bca{} is $\rho = -0.571$ ($p = 0.180$) for Llama 3.1 8B and $\rho = 0.0$ ($p = 1.0$) for Qwen 3.5 9B across the seven benchmark configurations. The mechanistic amplification pattern is consistent with scale-dependent late-layer commitment concentration, though cross-benchmark amplification-BCA correlations within individual model families are weak at this sample size.

\section{Faithfulness-Utility Scatter Plot}
\label{app:utility-scatter}

\begin{figure}[t]
\centering
\includegraphics[width=0.75\columnwidth]{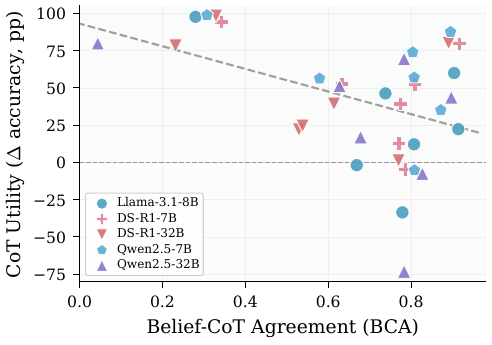}
\caption{\bca{} versus CoT utility ($\Delta$ accuracy, pp) across 35 configurations (5 models $\times$ 7 benchmarks). The Pearson correlation is significantly negative ($r = -0.42$, $p = 0.012$; bootstrap 95\% CI $[-0.64, -0.13]$). ProsQA (high-leverage cluster at low \bca{}, high utility) is interpretable as the clearest capability--oversight trade-off regime (see \Cref{app:utility-scatter}).}
\label{fig:bca-utility}
\vspace{-4mm}
\end{figure}

As shown in \Cref{fig:bca-utility}, across the 35 configurations with direct-answer baselines, mean direct accuracy is 27.0\%, mean CoT accuracy is 68.6\%, and mean CoT utility is +41.6 percentage points. The overall Pearson correlation between \bca{} and CoT utility is $r = -0.421$ ($p = 0.012$; bootstrap 95\% CI $[-0.64, -0.13]$, $B = 10{,}000$), with 4 configurations showing a CoT-hurts effect of at least five percentage points. All five within-model correlations are negative ($r \in [-0.52, -0.39]$), though individually non-significant at $n = 7$ per model (Llama-8B: $r = -0.52$; Qwen2.5-32B: $r = -0.47$; DS-R1-7B: $r = -0.44$; Qwen2.5-7B: $r = -0.44$; DS-R1-32B: $r = -0.39$).

\paragraph{Sensitivity to ProsQA.}
The Spearman rank correlation is directionally consistent ($\rho = -0.245$, $p = 0.155$) but not significant, indicating sensitivity to high-leverage points. A leave-one-benchmark-out analysis identifies ProsQA as the primary leverage point: removing it yields $r = +0.09$ ($p = 0.61$). In contrast, removing any other single benchmark preserves the negative direction: w/o PrOntoQA $r = -0.43$ ($p = 0.014$), w/o MMLU-Pro $r = -0.43$ ($p = 0.010$), w/o BBH-TSO $r = -0.41$ ($p = 0.017$), w/o MATH-500 $r = -0.32$ ($p = 0.078$), w/o BBH-LD $r = -0.32$ ($p = 0.069$), w/o GPQA-D $r = -0.26$ ($p = 0.145$). Three of six remaining benchmarks maintain significance at $p < 0.05$.

ProsQA's leverage is interpretable: it is the only graph-planning task where direct (no-CoT) accuracy is near zero across all models yet CoT accuracy exceeds 94\%, creating extreme CoT utility values ($\Delta > 0.78$) paired with uniformly low \bca{} ($< 0.35$). This is exactly the regime where the capability--oversight trade-off should be strongest: the model cannot solve graph-planning without explicit reasoning, so CoT is maximally useful, yet the logit lens reveals that the answer is resolved internally well before the trace converges. We therefore interpret ProsQA not as an outlier to be discarded but as the clearest empirical instantiation of the trade-off the paper identifies. Nonetheless, we caution that the overall correlation is moderate in strength and its significance depends on including tasks with extreme CoT gains.

\section{Probe Validation: Logit Lens and Activation Patching}

\subsection{Logit Lens Validation}
\label{app:logitlens}

We validate the logit lens through five checks. (1) \emph{Final-layer tautology}: the logit lens at the final layer matches the model's actual output token with $>$99\% accuracy. (2) \emph{Correctness conditioning}: \bca{} is significantly higher for correct instances (DS-R1: $p = 0.001$; Llama: $p = 0.037$), consistent with the lens capturing genuine answer-commitment information rather than noise. (3) \emph{Natural calibration}: on PrOntoQA (symbolic logic with deterministic step structure), Llama 3.1 8B achieves \bca{} = 0.914, indicating well-calibrated answer-commitment tracking on tasks with clear intermediate states. (4) \emph{Discriminative validity}: \bca{} spans the range 0.044--0.916 with systematic structure across models and benchmarks, ruling out ceiling/floor effects. (5) \emph{Patchscopes concordance}: we apply Patchscopes \citep{ghandeharioun2024patchscopes} to independently validate logit lens answer-commitment extraction across 3 models $\times$ 7 benchmarks (8,864 total steps). The logit lens agrees with Patchscopes on 75.6\% of steps at mid-depth layers and 74.7\% at final layers. Critically, the logit-lens-only disagreement rate is near-zero (0.0\% mid, 1.5\% final), meaning the logit lens almost never claims ground-truth agreement when Patchscopes disagrees. All substantive disagreements are Patchscopes-only (24.4\%), where Patchscopes detects additional faithfulness that the logit lens misses. This confirms that the logit lens is a \emph{conservative} estimator: any unfaithfulness it detects is validated by an independent method, and our reported \bca{} values are lower bounds on true answer-commitment alignment under this binary formulation.

\paragraph{Tuned lens concordance.} We additionally train affine probes per layer (tuned lens) on 30 calibration instances and compare the binary belief classification (believes ground truth vs.\ not) with the logit lens across 4 model-benchmark configurations. Agreement rates are: Qwen2.5-7B/MATH-500 99.2\%, Qwen2.5-7B/ProsQA 100\%, Llama 3.1 8B/MATH-500 99.8\%, DS-R1-7B/MATH-500 99.4\%. The near-perfect concordance ($\geq$99.2\% across all configurations) confirms that the linearity assumption underlying the logit lens does not introduce systematic bias into our belief extraction.

\subsection{Activation Patching Details}
\label{app:patching}

We apply direction ablation (\Cref{sec:causal-method}) to two matched pairs: Qwen2.5-7B-Instruct vs.\ DeepSeek-R1-Distill-Qwen-7B ($n = 60$ instances per configuration) and Qwen2.5-32B-Instruct vs.\ DeepSeek-R1-Distill-Qwen-32B ($n = 20$). Each pair is tested on three benchmarks (MATH-500, ProsQA, GPQA-Diamond) at four layers corresponding to approximately 50\%, 75\%, 86\%, and 93--97\% of model depth. For each instance, the answer-token direction $\hat{\mathbf{d}}_{a^*}$ is extracted as the normalized column of the unembedding matrix. At the target layer, this direction is projected out from the hidden state at the final token position with strength $\alpha \in \{1, 3, 5, 10\}$ (\Cref{eq:ablation}), and the model generates a new response (up to 256 tokens). We report the strength that produces the largest change in answer probability. We measure the \emph{correctness change rate}: whether the model's final answer switches between correct and incorrect after ablation.

\Cref{tab:patching-full} reports the full correctness change rate at each layer.

\begin{table}[h]
\centering
\small
\begin{tabular}{@{}llcccc@{}}
\toprule
\textbf{Model} & \textbf{Benchmark} & \textbf{50\%} & \textbf{75\%} & \textbf{86\%} & \textbf{93\%$^\dagger$} \\
\midrule
\multicolumn{6}{@{}l}{\emph{7B scale ($n = 60$; layers 14/21/24/26 of 28)}} \\
\midrule
Qwen2.5-7B & MATH-500 & 7\% & 7\% & 23\% & 18\% \\
Qwen2.5-7B & ProsQA & 8\% & 5\% & 8\% & 8\% \\
Qwen2.5-7B & GPQA-D & 7\% & 3\% & 2\% & 0\% \\
\midrule
DS-R1-7B & MATH-500 & 12\% & 5\% & 5\% & 10\% \\
DS-R1-7B & ProsQA & 38\% & 25\% & 27\% & 35\% \\
DS-R1-7B & GPQA-D & 5\% & 0\% & 0\% & 0\% \\
\midrule
\multicolumn{6}{@{}l}{\emph{32B scale ($n = 20$; layers 32/48/55/62 of 64)}} \\
\midrule
Qwen2.5-32B & MATH-500 & 25\% & 5\% & 10\% & 5\% \\
Qwen2.5-32B & ProsQA & 15\% & 10\% & 30\% & 25\% \\
Qwen2.5-32B & GPQA-D & 0\% & 0\% & 0\% & 0\% \\
\midrule
DS-R1-32B & MATH-500 & 40\% & 20\% & 10\% & 10\% \\
DS-R1-32B & ProsQA & 10\% & 25\% & \textbf{55\%} & 30\% \\
DS-R1-32B & GPQA-D & 10\% & 5\% & 10\% & 5\% \\
\bottomrule
\end{tabular}
\caption{Direction ablation correctness change rates by model, benchmark, and layer depth fraction. Bold = highest single cell. $^\dagger$93\% for 7B (layer 26/28); 97\% for 32B (layer 62/64).}
\label{tab:patching-full}
\end{table}

Key observations: (1) The strongest single effect is DS-R1-Distill-Qwen-32B on ProsQA at 86\% depth (55\% correctness change), confirming that detected answer-commitment signals are causally relevant at 32B scale. The layer gradient ($10\%{\rightarrow}25\%{\rightarrow}55\%{\rightarrow}30\%$) is consistent with late-layer concentration on this benchmark. (2) DS-R1-Distill-Qwen-7B shows uniformly high effects on ProsQA across all layers (25--38\%), consistent with heavy reliance on latent computation for graph-based reasoning (recall \bca{} $= 0.342$). (3) DS-R1-Distill-Qwen-32B on MATH-500 peaks at 50\% depth (40\%), indicating that the late-layer concentration pattern is task-dependent rather than universal. (4) Qwen2.5-7B on MATH-500 shows a late-layer peak (7\%$\rightarrow$7\%$\rightarrow$23\%$\rightarrow$18\%), mirroring the three-phase commitment pattern. (5) GPQA-Diamond shows minimal effects at both scales, likely because expert-level science questions involve distributed rather than concentrated answer representations. (6) Overall, the 32B pair (mean 16.7\% DS-R1, 11.7\% Qwen2.5) shows comparable causal effect magnitudes to the 7B pair (16.1\% DS-R1, 12.8\% Qwen2.5), confirming that the detected belief signals remain causally relevant at scale.

\begin{table}[t]
\centering
\small
\begin{tabular}{@{}lrrr@{}}
\toprule
\textbf{Benchmark} & \textbf{7B} & \textbf{14B} & \textbf{32B} \\
\midrule
MATH-500  & $+0.054$ & $+0.019$ & $-0.013$ \\
PrOntoQA  & $-0.036$ & $-0.001$ & $-0.149$ \\
ProsQA    & $+0.035$ & $+0.129$ & $+0.285$ \\
MMLU-Pro  & $+0.021$ & $+0.052$ & $-0.666$ \\
BBH-LD    & $-0.096$ & $-0.218$ & $-0.289$ \\
BBH-TSO   & $+0.006$ & $+0.388$ & $+0.107$ \\
GPQA-D    & $-0.022$ & $+0.023$ & $-0.014$ \\
\midrule
\textbf{Mean} & $-0.005$ & $+0.056$ & $-0.106$ \\
\bottomrule
\end{tabular}
\caption{Architecture-controlled pipeline effect on \bca{}, computed as DS-R1 minus Qwen2.5 at the same scale. Positive values indicate higher alignment after the reasoning pipeline; negative values indicate lower alignment.}
\label{tab:pipeline}
\vspace{-3mm}
\end{table}

\section{Human Annotation Validation}
\label{app:human-annotation}

To complement the LLM judge validation (\Cref{app:judges}), three human annotators independently labeled 100 stratified mismatch instances sampled across model families and benchmarks. Each annotator assigned one of the five taxonomy categories (CS, HR, CT, SEC, PC) to each instance based on the CoT text and the answer-commitment trajectory summary. Inter-annotator agreement reaches Gwet's $\mathrm{AC1} = 0.71$ on the five-way classification, comparable to the LLM judge agreement ($\mathrm{AC1} = 0.78$). The human labels confirm that the taxonomy categories correspond to qualitatively distinguishable phenomena: annotators most reliably identify Confabulated Steps (CS) and Premature Convergence (PC), with most disagreements occurring between CT and HR, consistent with the multi-label overlap documented in \Cref{app:multilabel}.

\section{Qualitative and Vacuousness Analysis}

\subsection{Qualitative Examples of Taxonomy Categories}
\label{app:qualitative}

We present representative examples of each taxonomy category to illustrate the qualitative differences among mismatch types. All examples are drawn from the current E1 evaluation set and are lightly edited for space.

\paragraph{Confabulated Steps (CS).}
DS-R1-Distill-Qwen-32B on ProsQA (instance \texttt{prosqa\_59}): The model's internal answer probability exceeds 0.8 from step~1, yet the trace continues for 30+ additional steps, generating text such as ``\emph{Let me verify by tracing the path: A connects to B, B connects to C\ldots}'' Each subsequent step induces $\Delta p_{\mathrm{ans}} < 0.01$, meeting the CS criterion. The reasoning is logically valid but temporally downstream of answer formation under our proxy.

\paragraph{Hidden Reasoning (HR).}
Llama 3.1 8B on MATH-500: Between steps~3 and~4, the answer probability jumps from 0.12 to 0.67 ($|\Delta p_{\mathrm{ans}}| = 0.55$), yet the CoT text at step~4 reads ``\emph{Substituting $x = 3$ into the equation\ldots}'' with no change in the explicit answer state. The large internal shift without corresponding explicit progress suggests that a significant inference occurred outside the visible trace.

\paragraph{Contradictory States (CT).}
Qwen 3.5 9B on GPQA-Diamond: At step~5, the answer-commitment proxy assigns probability 0.72 to answer (B), yet the trace explicitly endorses answer (C). This high-confidence disagreement ($\bca = 0$, probe confidence $= 0.72 > 0.5$) reflects a direct conflict between internal commitment and written reasoning.

\paragraph{Silent Error Correction (SEC).}
Qwen2.5-14B on BBH-Logical Deduction: The answer-commitment proxy disagrees with the trace for three consecutive steps (steps~4--6), then re-aligns at step~7 without any explicit correction or backtracking in the text. The model appears to have internally revised its answer and then resumed writing consistent reasoning without acknowledging the revision.

\paragraph{Premature Convergence (PC).}
DS-R1-Distill-Qwen-7B on MATH-500: The answer probability exceeds $\tau = 0.8$ at step~2 ($\ctg = 4$), but the trace does not state the final answer until step~6. The four intervening steps show decreasing $\Delta p_{\mathrm{ans}}$ values, consistent with the model writing out derivation steps after having already committed to a particular answer internally.

\subsection{Extended Qualitative Analysis}
\label{app:extended-qualitative}

We present additional qualitative examples that complement the single-instance illustrations in \Cref{app:qualitative} with multi-instance contrasts across model families.

\paragraph{Confabulated steps: surface plausibility vs.\ belief stasis.}
Three examples illustrate the contrast between logically valid reasoning and flat answer-commitment trajectories.
\textbf{(a)}~DS-R1-14B on MATH-500 (\texttt{math500\_4}): $p_{\mathrm{ans}} = 0.97$ from step~1; the model then computes average speeds for five runners across 10 additional steps ($\max |\Delta p_{\mathrm{ans}}| = 0.03$), each step arithmetically correct, arriving at ``\emph{Therefore, the student with the greatest average speed is Evelyn}.'' Every calculation is valid, yet all 10 post-commitment steps are temporally downstream of answer formation.
\textbf{(b)}~Llama 3.1 8B on GPQA-Diamond (\texttt{gpqa\_diamond\_110}): $p_{\mathrm{ans}} \geq 0.89$ on the correct answer (B) throughout 15 steps, yet the explicit calculation arrives at $v = 0.5$~m/s, contradicting the ground-truth answer of 1.58~m/s. The model then writes: ``\emph{However, I noticed that the solution provided does not match any of the answer choices.}'' The internal commitment and the explicit derivation are not merely desynchronized but substantively divergent---the model ``knows'' the answer while its visible reasoning computes a different value.
\textbf{(c)}~DS-R1-7B on BBH-LD (\texttt{bbh\_ld\_130}): a perfectly flat trajectory ($\Delta p_{\mathrm{ans}} = 0.000$) across six steps of logical deduction, all written in DS-R1's characteristic extended-thinking format.

\paragraph{Direction ablation: before and after.}
Projecting out the answer-token direction at layer~55 of DS-R1-32B on ProsQA produces three qualitatively distinct effects.
At $\alpha = 5$ (\texttt{prosqa\_59}), the model begins coherently (``\emph{Let's break down the problem step by step\ldots}'') but degenerates into repetitive fragments of the answer token (``\emph{Sallyumpusumpusumpusumpus\ldots}'')---a direct signature of the causal manipulation.
At $\alpha = 10$ (\texttt{prosqa\_33}), generation collapses entirely into the ablated token fragment (``\emph{ump\textbackslash numpump\textbackslash numpump\ldots}'').
At $\alpha = 10$ (\texttt{prosqa\_46}), the model preserves sentence structure but selectively truncates nonce-word answer tokens (``\emph{impu}'' for ``impus'', ``\emph{wumpu}'' for ``wumpus''), demonstrating that the ablated direction is localized to the answer representation while leaving syntax intact.
A complementary pattern occurs at 7B: on \texttt{prosqa\_26}, ablation \emph{corrects} a previously wrong answer (original incorrect $\to$ patched correct), confirming that the detected direction was causally responsible for the incorrect commitment rather than merely correlated with it.

\paragraph{Hidden reasoning: invisible inference.}
Qwen2.5-14B on MATH-500 (\texttt{math500\_289}): $p_{\mathrm{ans}}$ jumps from 0.00 to 1.00 between steps~2 and~3. At step~2 ($p = 0.00$) the model writes ``\emph{By Cauchy-Schwarz, $(a+b+c+d)(1/a+1/b+1/c+1/d) \geq (1+1+1+1)^2 = 16$.}''---the genuine derivation. At step~3 ($p = 1.00$) it writes ``\emph{Therefore, the minimum value is 16.}'' The internal commitment crystallizes only when the answer is explicitly stated, not during the step that contains the actual inference. This HR pattern---reasoning visible, commitment invisible---is the qualitative inverse of CS and suggests that for mathematical derivations, the logit lens may capture \emph{answer-token} commitment rather than \emph{solution knowledge}, consistent with the proxy limitation noted in \Cref{sec:limitations}.

\subsection{Confabulated-Step Vacuousness Analysis}
\label{app:cs-vacuousness}

A natural objection to the ``confabulated steps'' interpretation is that the model may be performing genuine verification that happens not to change $p_{\mathrm{ans}}$. To test this, we analyze all instances with detected CS steps across 56 model-benchmark configurations using the cached answer-probability trajectories (a representative subset of the full evaluation set). For each instance, we check whether the committed answer token ever changes during CS steps, whether $p_{\mathrm{ans}}$ remains above the commitment threshold $\tau = 0.3$ throughout, and what fraction of CS steps show near-zero belief change.

Across 314 instances containing 1{,}704 CS steps:
\begin{itemize}[nosep]
    \item \textbf{100\%} of instances maintained stable answer commitment throughout all CS steps---the committed answer never changed.
    \item \textbf{90.1\%} of individual CS steps showed near-zero belief change ($\Delta p_{\mathrm{ans}} < 0.001$).
    \item \textbf{66.3\%} of CS steps had strong commitment ($p_{\mathrm{ans}} > 0.8$); the mean $p_{\mathrm{ans}}$ across all CS steps was 0.851.
    \item On average, CS steps constituted 30.9\% of total steps per instance.
\end{itemize}

The 100\% answer-stability rate across all 314 instances is the key finding: if CS steps were performing genuine verification, we would expect at least some instances where the verification process leads to an answer revision (a dip or change in $p_{\mathrm{ans}}$). Instead, the answer-commitment signal is entirely flat, consistent with vacuous continuation rather than active checking. Per-model rates of strong commitment range from 56.1\% (Qwen2.5-14B) to 81.8\% (Qwen2.5-7B), with all models showing 100\% stable argmax.

\section{Additional Sensitivity and Comparison Analyses}

\subsection{Cross-Method Faithfulness Comparison}
\label{app:cross-method}

To assess whether \bca{} captures information distinct from simpler faithfulness proxies, we compute an \emph{early commitment rate} (ECR) for each model-benchmark configuration: the fraction of instances where $p_{\mathrm{ans}} > \tau$ by the midpoint of the reasoning chain. ECR is analogous to the early-answering tests of \citet{lanham2023measuring}, computed from internal answer-probability trajectories rather than behavioral perturbation.

Across 56 configurations, the Spearman correlation between \bca{} and ECR is $\rho = 0.048$ ($p = 0.73$), indicating that the two metrics capture essentially orthogonal information. This is expected: ECR measures whether the model has committed early in absolute terms, while \bca{} measures the \emph{timing alignment} between latent commitment and explicit answer arrival across the full chain. A model can have high ECR (commits early) but also high \bca{} (if the CoT also converges early), or low ECR (commits late) with low \bca{} (if the CoT converges at a different time).

ECR shows no significant relationship with CoT utility ($\rho = -0.12$, $p = 0.69$, $n = 14$), paralleling the \bca{}--utility null result. The near-zero \bca{}--ECR correlation confirms that \bca{} provides a measurement lens that is complementary to, rather than redundant with, response-level early-answering metrics.

\subsection{Per-Scale Taxonomy Shift Details}
\label{app:taxonomy-shift}

We report the full per-scale chi-squared analysis of taxonomy shifts between matched Qwen2.5 (base) and DS-R1 (pipeline-trained) pairs. At 7B ($N_{\text{base}}\!=\!17{,}338$; $N_{\text{pipe}}\!=\!17{,}971$), the taxonomy is largely preserved ($\chi^2 = 119.8$, $p < 10^{-24}$, Cram\'{e}r's $V = 0.058$): confabulated steps account for 57.9\% versus 53.0\%, with all other categories shifting by less than 3.5~pp. At 14B ($N_{\text{base}}\!=\!36{,}025$; $N_{\text{pipe}}\!=\!21{,}881$), the effect grows ($\chi^2 = 1{,}399$, $V = 0.155$) and the pipeline reduces average mismatches per instance from 24.2 to 14.7 (a 39\% reduction), consistent with its positive \bca{} effect. At 32B ($N_{\text{base}}\!=\!17{,}538$; $N_{\text{pipe}}\!=\!25{,}858$), the shift is largest ($\chi^2 = 3{,}629$, $V = 0.289$): confabulated steps drop from 57.6\% ($N\!=\!10{,}096$) to 37.4\% ($N\!=\!9{,}674$; $-20.2$~pp), while contradictory states surge from 7.1\% ($N\!=\!1{,}246$) to 28.9\% ($N\!=\!7{,}480$; $+21.8$~pp). The 7B-vs-32B aggregate \bca{} contrast does not reach significance (paired Wilcoxon: $p = 0.578$), and the 32B mean drops to $-0.012$ when the large MMLU-Pro outlier ($\Delta = -0.666$) is excluded. The mechanistic evidence in \Cref{app:mechanistic} suggests that at 32B, the pipeline concentrates answer commitment into late layers whose representations are less tightly coupled to the generated text, creating opportunities for the trace to endorse a different answer than the one the model has internally resolved.

\subsection{BCA--Utility Sensitivity Analysis}
\label{app:utility-sensitivity}

The Spearman rank correlation between \bca{} and CoT utility is directionally consistent but not significant ($\rho = -0.25$, $p = 0.16$), indicating sensitivity to high-leverage points. A leave-one-benchmark-out analysis confirms that ProsQA is the primary leverage point: removing it yields $r = +0.09$ ($p = 0.61$), whereas removing any other single benchmark preserves the negative direction ($r \in [-0.43, -0.26]$) with three of six remaining benchmarks maintaining significance ($p < 0.05$). ProsQA's leverage is interpretable rather than artifactual: it is the only graph-planning task where direct accuracy is near zero yet CoT accuracy exceeds 94\%, making it the clearest example of CoT enabling capability that the model cannot achieve without explicit reasoning---precisely the regime where the capability--oversight trade-off should be strongest. All five within-model correlations are negative ($r \in [-0.52, -0.39]$), though individually non-significant at $n = 7$, providing directional consistency across model families.

\section{Temperature Sensitivity}
\label{app:temperature}

All main experiments use greedy decoding for reproducibility, but reasoning models are typically deployed with temperature sampling. To check whether the qualitative patterns hold under sampling, we run two key configurations at $T = 0.6$ (top-$p = 0.95$) with three samples each ($n = 60$ instances per configuration).

\begin{table}[h]
\centering
\small
\begin{tabular}{@{}llccc@{}}
\toprule
\textbf{Config} & \textbf{Decoding} & \textbf{Accuracy} & \textbf{CoT Length} & \textbf{Consistency} \\
\midrule
\multirow{2}{*}{Llama 3.1 8B / MATH} & Greedy & 0.667 & 2{,}043 & --- \\
& $T\!=\!0.6$ & 0.600 $\pm$ 0.014 & 2{,}445 & 0.75 \\
\midrule
\multirow{2}{*}{DS-R1-7B / ProsQA} & Greedy & 0.950 & 2{,}800 & --- \\
& $T\!=\!0.6$ & 0.972 $\pm$ 0.016 & 2{,}724 & 0.93 \\
\bottomrule
\end{tabular}
\caption{Temperature sensitivity check. Accuracy and CoT length under greedy vs.\ temperature sampling. Consistency = fraction of instances where all three samples produce the same answer.}
\label{tab:temperature}
\end{table}

\Cref{tab:temperature} summarizes the results. Two observations support the robustness of our main findings. First, accuracy under temperature sampling is within 7 percentage points of greedy in both configurations, indicating that temperature does not fundamentally alter the model's reasoning capacity. Second, CoT traces are 15--20\% longer under sampling for Llama 3.1 8B yet accuracy drops slightly, consistent with the main-text finding that additional reasoning text does not necessarily reflect additional inferential contribution. For DS-R1-7B on ProsQA (greedy \bca{} = 0.342), accuracy is comparably high under sampling (0.972), and 93\% of instances produce consistent answers across all three samples, reinforcing that the model has committed to its answer before the extended reasoning trace concludes. These results suggest that the performative-reasoning pattern identified under greedy decoding is not an artifact of the decoding strategy.

\section{Detailed Differentiation from Closely Related Work}
\label{app:differentiation}

Several concurrent works share parts of our motivation---especially the observation that a model can commit to an answer internally before its trace reveals it---so we briefly make the differences explicit. \Cref{tab:differentiation} compares \dcc{} against ten closely related works along the axes we view as most consequential for diagnosing CoT unfaithfulness: measurement granularity, internal probe family, classification granularity, architecture-matched pipeline control, and capability--oversight quantification.

\begin{table*}[h]
\centering
\scriptsize
\setlength{\tabcolsep}{4pt}
\renewcommand{\arraystretch}{1.15}
\resizebox{\textwidth}{!}{%
\begin{tabular}{@{}lllllc@{}}
\toprule
\textbf{Work} & \textbf{Granularity} & \textbf{Internal probe} & \textbf{Classification} & \textbf{Arch-matched} & \textbf{Cap--Over.} \\
\midrule
\citet{lanham2023measuring}        & Response            & Truncation / perturb.          & Binary                   & No    & $\times$ \\
\citet{turpin2024language}         & Response            & Bias injection                 & Binary                   & No    & $\times$ \\
\citet{chen2025reasoning}          & Response            & Hint injection                 & Binary                   & No    & $\times$ \\
\citet{arcuschin2025chain}         & Response            & Contrastive prompting          & Binary                   & No    & $\times$ \\
\citet{tutek2025fur}               & Step                & Parameter unlearning           & Scalar drop              & No    & $\times$ \\
\citet{ye2026faithfulness_decay}   & Step (NLDD)         & Corruption + conf.\ diff.      & Scalar decay             & No    & $\times$ \\
\citet{young2026lietome,young2026whymodelsknow} & Response / token & Hint injection     & Binary                   & No    & $\times$ \\
\citet{palod2025performative_thinking} & Trace length    & None (behavioral)              & Length--complexity corr. & No    & $\times$ \\
\citet{boppana2026reasoning_theater} & Token / step      & Trained attn.\ probes          & Binary (perf.)           & No    & $\times$ \\
\citet{hase2026cst}                & Training-time       & Counterfactual simulation      & N/A                      & No    & $\times$ \\
\midrule
\rowcolor{gray!10} \dcc{} (ours)   & \textbf{Step (timing)} & \textbf{Logit lens + Patchscopes + tuned lens} & \textbf{Five-way} & \textbf{Yes (3 scales)} & $\checkmark$~($r{=}{-}0.42$) \\
\bottomrule
\end{tabular}%
}
\caption{Axis-by-axis comparison of \dcc{} against closely related work. ``Arch-matched'' indicates whether the work uses architecture-controlled model pairs to isolate training-pipeline effects; ``Cap--Over.'' indicates whether the work quantifies a capability--oversight trade-off. \dcc{} is the only framework combining a training-free multi-method probing pipeline, a five-way step-level taxonomy, architecture-matched scale comparison, and an explicit capability--oversight correlation.}
\label{tab:differentiation}
\end{table*}

\paragraph{What makes \dcc{} different.}
\dcc{} is distinguished from prior work on four concrete points. \textbf{(i) Timing rather than causal importance.} Step-level methods such as \citet{tutek2025fur} and \citet{ye2026faithfulness_decay} quantify how much each step \emph{matters causally} via unlearning or corruption; \dcc{} instead measures \emph{when} latent answer commitment stabilizes relative to the trace, which is why \bca{} is sensitive to post-commitment continuation that causal-importance metrics can score as inert. \textbf{(ii) Training-free multi-lens probing.} Related work splits here: \citet{palod2025performative_thinking} questions whether longer traces reflect more adaptive computation, while \citet{boppana2026reasoning_theater} studies latent belief/trace mismatch with trained attention probes whose accuracy is highly sensitive to probe architecture (87.98\% vs.\ 31.85\% on MMLU in \citealt{boppana2026reasoning_theater}); \dcc{} uses the logit lens cross-validated with Patchscopes, tuned lens, and causal direction ablation (\Cref{app:patching}), all training-free. \textbf{(iii) Five-way taxonomy instead of a binary verdict.} Performative-vs-genuine framings collapse qualitatively different failure modes: at 32B the R1 pipeline reduces confabulated steps by 20~pp while contradictory states \emph{grow} by 22~pp (\Cref{sec:results-pipeline})---a swap a binary framing would score as roughly unchanged, but that changes surface plausibility and oversight difficulty substantially. \textbf{(iv) Architecture-matched pipeline control with an explicit capability--oversight correlation.} No prior work in \Cref{tab:differentiation} pairs a reasoning-trained model to its backbone; cross-family comparisons (e.g.\ DeepSeek-R1-671B vs.\ GPT-OSS-120B in \citealt{boppana2026reasoning_theater}) confound architecture with training. Our Qwen2.5 vs.\ DeepSeek-R1-Distill pairs at 7B/14B/32B hold the backbone fixed, and we additionally report a Pearson $r{=}{-}0.42$ ($p{=}0.012$, 95\% CI $[-0.64,-0.13]$) between \bca{} and CoT utility across 35 configurations, turning the faithfulness--capability trade-off from an anecdote into a quantitative claim. These directions are complementary to the probes enabling early exit in \citet{boppana2026reasoning_theater} and the training recipe of \citet{hase2026cst}, and a natural extension of \dcc{} is to use \bca{} as a training-time monitor for faithfulness-improving interventions.

\section{Pure-Subgroup Causal Truncation}
\label{app:truncation-v3}

Section~\ref{sec:results-taxonomy} reports five mismatch categories as a descriptive decomposition. To test whether each category has a \emph{step-level causal signature}, we run a within-instance paired-truncation experiment restricted to instances where exactly one detector fires above severity $0.30$ (``pure'' subgroups). Restricting to pure instances is required because the five detectors are structurally correlated (\Cref{app:multilabel}), and pooling over multi-label instances mixes opposite-sign effects that cancel in the marginal.

\paragraph{Within-instance paired design.}
For each pure-subgroup instance $i$ where category $c$ fires, we pick a \emph{characteristic firing step} $S_i^c$ (highest-$p_{\mathrm{ans}}$ CS firing; largest-$|\Delta p_{\mathrm{ans}}|$ HR jump; most-confident CT disagreement), truncate the CoT at $S_i^c$, append a reasoning-model-aware answer-elicitation suffix, and regenerate greedy. As paired controls on the \emph{same} instance we cut at $S_i^c + \delta$ with $\delta$ drawn from $\{-2,-1,1,2\}$ (neighbor control) and at a uniformly random step $R_i$ (uniform control). The paired causal statistics are
\[
\Delta^c_{i,\text{N}} = \mathbb{1}[\text{correct}(S_i^c)] - \mathbb{1}[\text{correct}(S_i^c + \delta)], \qquad
\Delta^c_{i,\text{U}} = \mathbb{1}[\text{correct}(S_i^c)] - \mathbb{1}[\text{correct}(R_i)].
\]
Because each control is on the same instance, instance-level difficulty and depth effects cancel out. The experiment runs on all 56 model-benchmark configurations used in \Cref{app:multilabel} (8 open-weight models $\times$ 7 benchmarks), pooling 2{,}088 pure-CS, 410 pure-HR, and 38 pure-CT paired triples.

\paragraph{Pooled results.}
\Cref{tab:truncation-v3-overall} reports the overall pooled paired $\Delta$ with paired $t$-tests and bootstrap 95\% CIs. Both CS and HR show statistically significant \emph{positive} deltas---cutting at a CS or HR firing \emph{helps} accuracy relative to either control. The effect size is much larger against a uniformly random cut ($\Delta_{\text{U}}$) than against a matched neighbor cut ($\Delta_{\text{N}}$), because the characteristic-step picker deliberately targets steps that matter for the detector's semantic signal, which is not true of the random control. CT shows a directional negative signal (cutting hurts, consistent with a load-bearing interpretation) but is underpowered at $n=38$.

\begin{table}[h]
\centering
\small
\begin{tabular}{lrrrrrr}
\toprule
Category & $n$ & $\Delta_{\text{vs N}}$ & 95\% CI & $p$ & $\Delta_{\text{vs U}}$ & $p$ \\
\midrule
CS (pure) & $2{,}088$ & $+0.028$ & $[+0.009, +0.047]$ & $0.0043$ & $+0.078$ & $<\!10^{-16}$ \\
HR (pure) & $410$     & $+0.046$ & $[+0.010, +0.085]$ & $0.0165$ & $+0.161$ & $<\!10^{-16}$ \\
CT (pure) & $38$      & $-0.053$ & $[-0.158, +0.053]$ & $0.3238$ & $-0.053$ & $0.160$ \\
\bottomrule
\end{tabular}
\caption{v3 pure-subgroup paired truncation, pooled over 56 model-benchmark configurations. Confabulated Steps (CS) show a significant \emph{positive} $\Delta$ vs both neighbor ($\delta=\pm 2$ step) and uniform-random controls, confirming CS vacuousness. HR also shows significant positive $\Delta$, in the direction opposite our original ``load-bearing hidden computation'' prediction. CT is directionally negative (load-bearing) but underpowered.}
\label{tab:truncation-v3-overall}
\end{table}

\paragraph{Task-type heterogeneity.}
The CS vacuousness effect is strongly task-dependent (\Cref{tab:truncation-v3-benchmark}): it is large and significant on simple-answer benchmarks (BBH-LD $\Delta{=}{+}0.080$, BBH-TSO $\Delta{=}{+}0.077$, MMLU-P $\Delta{=}{+}0.045$, all $p<0.02$; ProsQA $\Delta{=}{+}0.103$, $p{=}0.044$), null on heavy-computation benchmarks (MATH-500 $\Delta{=}{-}0.003$, GPQA-D $\Delta{=}{-}0.006$), and trends negative on PrOntoQA ($\Delta{=}{-}0.045$, $p{=}0.10$). The interpretation is that ``vacuous post-commitment continuation'' accurately describes late-stage CoT on tasks whose answer is determined early and the remaining steps are rhetorical padding, but on tasks whose answer is still being computed at the CS step the cut removes meaningful computation rather than padding.

\begin{table}[h]
\centering
\small
\begin{tabular}{lrrrr}
\toprule
Benchmark & $n$ & $\Delta_{\text{vs N}}$ & 95\% CI & $p$ \\
\midrule
BBH Logical Deduction & $400$ & $+0.080$ & $[+0.025, +0.135]$ & $0.0036$ \\
BBH TSO               & $325$ & $+0.077$ & $[+0.028, +0.129]$ & $0.0025$ \\
ProsQA                & $39$  & $+0.103$ & $[+0.026, +0.205]$ & $0.0439$ \\
MMLU-Pro              & $400$ & $+0.045$ & $[+0.018, +0.075]$ & $0.0034$ \\
GPQA-Diamond          & $161$ & $-0.006$ & $[-0.043, +0.031]$ & $0.7400$ \\
MATH-500              & $363$ & $-0.003$ & $[-0.044, +0.041]$ & $0.8999$ \\
PrOntoQA              & $400$ & $-0.045$ & $[-0.098, +0.008]$ & $0.1032$ \\
\bottomrule
\end{tabular}
\caption{Per-benchmark pure-CS paired $\Delta_{\text{vs N}}$. The CS vacuousness effect is positive on simple-answer tasks and null or negative on computation-heavy or rare-token tasks.}
\label{tab:truncation-v3-benchmark}
\end{table}

\paragraph{Scale dependence.}
Aggregating by model parameter scale (\Cref{tab:truncation-v3-scale}), the CS vacuousness effect peaks at 12--14B ($\Delta{=}{+}0.056$, $p{=}5{\times}10^{-4}$), is near-null at 7--9B, and near-null at 32B. HR, by contrast, shows its strongest positive $\Delta$ at 32B ($\Delta{=}{+}0.101$, $p{=}0.011$). A possible reading: at 12--14B the model commits confidently enough to generate detectable vacuous tails but is not yet sophisticated enough to use post-commitment steps for verification; at 32B post-commitment continuation becomes more selective, so the pure-CS causal signal washes out while HR-firing steps at the same scale turn out to be verification-adjacent rather than forward-computing.

\begin{table}[h]
\centering
\small
\begin{tabular}{lrrrrrr}
\toprule
Scale & $n_{\text{CS}}$ & $\Delta_{\text{CS}}$ & $p_{\text{CS}}$ & $n_{\text{HR}}$ & $\Delta_{\text{HR}}$ & $p_{\text{HR}}$ \\
\midrule
7--9B   & $841$ & $+0.008$ & $0.591$ & $178$ & $+0.039$ & $0.162$ \\
12--14B & $736$ & $+0.056$ & $\mathbf{5.3{\times}10^{-4}}$ & $153$ & $+0.026$ & $0.452$ \\
32B     & $511$ & $+0.022$ & $0.310$ & $79$  & $+0.101$ & $\mathbf{0.011}$ \\
\bottomrule
\end{tabular}
\caption{Pure-subgroup $\Delta_{\text{vs N}}$ by parameter-scale band. CS vacuousness peaks at 12--14B; HR vacuousness peaks at 32B.}
\label{tab:truncation-v3-scale}
\end{table}

\paragraph{Interpretation.}
Three substantive implications follow. \emph{(i) CS is causally vacuous on simple-answer tasks.} The within-instance pure-subgroup $\Delta_{\text{vs U}} = +0.078$ ($p<10^{-16}$) confirms that cutting through a CS-firing step is a strictly better choice than cutting at a uniformly random step, and the $+0.080$ effect on BBH Logical Deduction alone is the cleanest validation the paper offers of CS as ``vacuous post-commitment continuation.'' \emph{(ii) HR is not universally load-bearing.} Our original prediction---that HR firings (``hidden computation'' jumps) would be causally load-bearing and therefore show $\Delta < 0$---is contradicted at the pooled level ($\Delta_{\text{vs N}} = +0.046$, $p=0.016$) and even more strongly against the uniform control ($\Delta_{\text{vs U}} = +0.161$, $p<10^{-16}$). One natural re-reading is that large $|\Delta p_{\mathrm{ans}}|$ jumps flag steps \emph{immediately after} which the answer has been locked in, so cutting at the pre-jump step still leaves a prefix sufficient for the model to re-derive the committed answer via greedy decoding from the neighboring context. \emph{(iii) The five-way taxonomy is a descriptive decomposition.} Combined with the structural non-isolability of PC and SEC (\Cref{app:multilabel}), the pure-CS/pure-HR results support reading the taxonomy as a descriptive breakdown of correlated facets of trace unfaithfulness rather than five independent causal mechanisms. This is compatible with the paper's original claim that \dcc{} measures timing alignment (\bca{}) rather than causal step-importance.

\paragraph{Robustness: HR post-jump variant.}
The HR-surprise reading above---that a large $|\Delta p_{\mathrm{ans}}|$ jump marks the step \emph{after} which the answer has been locked in---makes a direct prediction: cutting just \emph{after} the jump should be at least as vacuous as cutting just before it. We test this by rerunning the paired truncation pipeline with the HR treatment step shifted from the pre-jump position (step\_range$[0]$) to the post-jump position (step\_range$[1]+1$) on 49 of the 56 main configurations (DS-R1-32B is excluded because its model-path symlink became invalid between the E3 main run and this robustness run). CS and CT pickers are held identical to the main run so that those categories serve as internal sanity checks. The clean test is a within-instance paired comparison on the subset of HR instances where both pre-jump and post-jump treatment steps are defined ($N=73$). On this paired subset, the pre-jump cut gives $\Delta_{\text{vs N}} = -0.027$ ($p=0.60$, not significant), while the post-jump cut gives $\Delta_{\text{vs N}} = +0.137$ ($p=0.011$); a within-instance paired $t$-test on the difference (post minus pre) yields $+0.164$ ($p=0.027$). Both cut positions lie in the same (vacuous) direction when pooled across all instances where each is applicable, but the post-jump cut is markedly larger in magnitude on the paired subset. This supports the re-reading: cutting at the pre-jump step leaves a prefix sufficient for greedy decoding to re-derive the committed answer, whereas cutting at the post-jump step preserves the commit moment and strips the now-redundant post-commit tail. The main conclusion---that HR is causally vacuous on average in pure subgroups---is robust to the exact choice of step within the HR firing window, and in fact becomes stronger when the HR cut is placed at its semantically more interpretable position.

\paragraph{Complementary test: CS content is causally inert under donor corruption.}
\label{app:cs-corruption}
To directly test whether CS steps are being \emph{read} during the original forward pass (as opposed to merely being unnecessary under regeneration), we run a corruption experiment that keeps the full CoT intact but replaces the tokens at the CS firing step with length-matched tokens from another instance's CS firing step (cross-instance donor swap). For each of $N = 1{,}733$ pure-CS instances pooled over 51 configurations, we compare corruption at three locations on the same instance: the CS firing step itself, a $\pm 2$ neighbor step, and a uniformly-random non-CS step. We measure the intervention two ways: Variant A reads the last-step answer probability $p(\text{ans} \mid \text{corrupted context})$ under forced decoding (logit-level); Variant B greedy-regenerates a short answer after an elicitation suffix and scores correctness (behavior-level).
Under Variant B, corrupting the CS step changes final-answer accuracy by $\Delta_{\text{CS}} = -0.002$ ($p = 0.78$, paired $t$-test, essentially a null), while corruption at a neighbor or uniformly-random step changes accuracy by $\Delta_{\text{N}} = -0.020$ ($p = 0.013$) and $\Delta_{\text{U}} = -0.021$ ($p = 0.007$) respectively. The within-instance paired contrasts are $\Delta_{\text{CS}} - \Delta_{\text{N}} = +0.017$ ($p = 0.039$) and $\Delta_{\text{CS}} - \Delta_{\text{U}} = +0.019$ ($p = 0.028$), both significant at the $p<0.05$ level: corrupting CS is measurably less damaging than corrupting any other position on the same instance. The effect concentrates at the 12--14B parameter band (paired $\Delta_{\text{CS}} - \Delta_{\text{N}} = +0.046$, $p = 0.003$) and on BBH-Tracking in particular ($\Delta_{\text{CS}} = +0.097$, $p = 6.1 \times 10^{-4}$). Variant A's logit-level signal is smaller because most pure-CS instances already have $p(\text{ans})$ close to 1.0 in the original context and saturate; the paired contrast nonetheless points in the same direction ($\Delta_{\text{CS}} - \Delta_{\text{N}} = -0.009$, $p = 0.004$ for the $\pm 2$ neighbor control, where a more negative value indicates less logit damage from CS corruption). Two observations follow. First, the CS step's tokens are not merely dispensable under re-generation (the truncation test above): they are actively ignored during the original forward pass, since replacing them with unrelated text from another CoT produces no measurable accuracy loss while the same replacement elsewhere in the trace does. Second, the specificity holds even though both neighbor and uniform corruption locations often also land within the post-commitment tail; the residual effect size at other-location corruption shows that the CS picker is selecting the step whose content contributes least to downstream generation. PrOntoQA is again the one benchmark where CS corruption \emph{hurts} accuracy ($\Delta_{\text{CS}} = -0.088$, $p = 4.6 \times 10^{-5}$), consistent with its outlier behavior in the main truncation test on rare-token graph reasoning with tight cross-step dependencies.

\paragraph{Limitations of this analysis.}
The pure-CT subgroup ($n=38$ across 56 configurations) is too small to reach $p<0.05$ individually despite the consistent load-bearing sign, and we report CT as directionally suggestive rather than confirmed. The CS and HR pure-subgroup selection necessarily excludes instances where a category co-fires with any other, which means the reported $\Delta$ does not estimate the causal effect of that category in its more common multi-label context; that latter estimand is not identifiable under the current detector definitions because the pure pool is sampled from the tail of a correlated joint distribution. Our findings should be read as bounds on the causally clean component of each category.

\newpage
\section*{NeurIPS Paper Checklist}

\begin{enumerate}

\item {\bf Claims}
    \item[] Question: Do the main claims made in the abstract and introduction accurately reflect the paper's contributions and scope?
    \item[] Answer: \answerYes{} 
    \item[] Justification: The abstract and introduction state a step-level empirical test of whether CoT traces remain temporally synchronized with internal answer commitment, and the listed contributions match the reported BCA, taxonomy, matched-pipeline, and CoT-utility analyses. The scope is limited to empirical evidence from the evaluated models and benchmarks rather than a universal claim about all models.

\item {\bf Limitations}
    \item[] Question: Does the paper discuss the limitations of the work performed by the authors?
    \item[] Answer: \answerYes{} 
    \item[] Justification: Section 7 discusses the answer-commitment proxy, possible nonlinear or multi-token beliefs, intervention-strength selection, overlapping taxonomy categories, the Qwen2.5/DeepSeek-R1 comparison scope, and the restriction to open-weight 7B--32B models.

\item {\bf Theory assumptions and proofs}
    \item[] Question: For each theoretical result, does the paper provide the full set of assumptions and a complete (and correct) proof?
    \item[] Answer: \answerNA{} 
    \item[] Justification: The paper does not present formal theoretical results, theorems, or proofs. The equations define operational measurements and interventions used in the empirical framework.

    \item {\bf Experimental result reproducibility}
    \item[] Question: Does the paper fully disclose all the information needed to reproduce the main experimental results of the paper to the extent that it affects the main claims and/or conclusions of the paper (regardless of whether the code and data are provided or not)?
    \item[] Answer: \answerYes{} 
    \item[] Justification: The paper discloses the models, benchmarks, inference setup, step parser, answer readout, BCA/CTG metrics, thresholds, statistical procedures, validation checks, and intervention protocol needed to reproduce the main empirical results. The experiments use existing open-weight models and public benchmarks rather than training a new model.

\item {\bf Open access to data and code}
    \item[] Question: Does the paper provide open access to the data and code, with sufficient instructions to faithfully reproduce the main experimental results, as described in supplemental material?
    \item[] Answer: \answerNo{} 
    \item[] Justification: At submission time, we do not provide an anonymized public code/data release with exact reproduction commands. The paper nevertheless describes the methodology and relies on public models and benchmarks; code and generated artifacts can be released after anonymity and licensing review.

\item {\bf Experimental setting/details}
    \item[] Question: Does the paper specify all the training and test details (e.g., data splits, hyperparameters, how they were chosen, type of optimizer) necessary to understand the results?
    \item[] Answer: \answerYes{} 
    \item[] Justification: Section 3 specifies the evaluated models, benchmarks, instance counts, greedy decoding setup, step segmentation procedure, hidden-state readout, BCA thresholding, confidence-interval construction, and compute budget. Since no models are trained in this work, optimizer and training hyperparameters are not applicable.

\item {\bf Experiment statistical significance}
    \item[] Question: Does the paper report error bars suitably and correctly defined or other appropriate information about the statistical significance of the experiments?
    \item[] Answer: \answerYes{} 
    \item[] Justification: The paper reports 95\% bootstrap confidence intervals with Bonferroni correction and appropriate statistical tests, including ANOVA, Bayesian model comparison, correlations, Wilcoxon tests, chi-squared tests with Cramer's V, and intervention/truncation significance tests. Sensitivity analyses for threshold choice and validation checks are reported in the appendix.

\item {\bf Experiments compute resources}
    \item[] Question: For each experiment, does the paper provide sufficient information on the computer resources (type of compute workers, memory, time of execution) needed to reproduce the experiments?
    \item[] Answer: \answerNo{} 
    \item[] Justification: The draft reports aggregate compute for the full evaluation, approximately 200 GPU-hours on four A100 GPUs, but does not yet provide per-experiment runtime, memory, and storage requirements. We therefore conservatively answer no and will add a more detailed compute table in the appendix.
    
\item {\bf Code of ethics}
    \item[] Question: Does the research conducted in the paper conform, in every respect, with the NeurIPS Code of Ethics \url{https://neurips.cc/public/EthicsGuidelines}?
    \item[] Answer: \answerYes{} 
    \item[] Justification: The authors have reviewed the NeurIPS Code of Ethics and the research is intended to conform to it. The work uses public benchmarks and existing models, does not collect private personal data, and does not release a high-risk model or scraped dataset.

\item {\bf Broader impacts}
    \item[] Question: Does the paper discuss both potential positive societal impacts and negative societal impacts of the work performed?
    \item[] Answer: \answerYes{} 
    \item[] Justification: The paper discusses positive implications for improving CoT-based oversight and process monitoring, while also identifying the negative risk that fluent CoT traces and step-level monitors may be misleading after internal answer commitment. The work is diagnostic rather than a deployment or capability-release paper.
    
\item {\bf Safeguards}
    \item[] Question: Does the paper describe safeguards that have been put in place for responsible release of data or models that have a high risk for misuse (e.g., pre-trained language models, image generators, or scraped datasets)?
    \item[] Answer: \answerNA{} 
    \item[] Justification: The paper does not release a pretrained language model, image generator, scraped dataset, or other high-risk asset requiring controlled release safeguards. The experiments use existing public models and benchmarks and report diagnostic measurements.

\item {\bf Licenses for existing assets}
    \item[] Question: Are the creators or original owners of assets (e.g., code, data, models), used in the paper, properly credited and are the license and terms of use explicitly mentioned and properly respected?
    \item[] Answer: \answerNo{} 
    \item[] Justification: The paper cites the original model and benchmark sources used in the experiments, but the current draft does not yet include an explicit asset table listing licenses and terms of use for each existing model and dataset. We therefore conservatively answer no and will add license and usage-term information in the appendix.

\item {\bf New assets}
    \item[] Question: Are new assets introduced in the paper well documented and is the documentation provided alongside the assets?
    \item[] Answer: \answerNA{} 
    \item[] Justification: The current submission does not release a new dataset, model, or packaged code asset. If code, generated traces, cached activations, or annotation files are released later, they should be documented with a README, license, intended-use statement, and reproduction instructions.

\item {\bf Crowdsourcing and research with human subjects}
    \item[] Question: For crowdsourcing experiments and research with human subjects, does the paper include the full text of instructions given to participants and screenshots, if applicable, as well as details about compensation (if any)? 
    \item[] Answer: \answerNA{} 
    \item[] Justification: Appendix K reports a small human annotation validation, but  all annotators were authors and no external participants were recruited, this item should be N/A.

\item {\bf Institutional review board (IRB) approvals or equivalent for research with human subjects}
    \item[] Question: Does the paper describe potential risks incurred by study participants, whether such risks were disclosed to the subjects, and whether Institutional Review Board (IRB) approvals (or an equivalent approval/review based on the requirements of your country or institution) were obtained?
    \item[] Answer: \answerNo{} 
    \item[] Justification: The current draft does not state an IRB or exemption status, participant-risk disclosure, or equivalent review procedure for the human annotation validation. We therefore conservatively answer no because the annotation study is mentioned but review status is not described.

\item {\bf Declaration of LLM usage}
    \item[] Question: Does the paper describe the usage of LLMs if it is an important, original, or non-standard component of the core methods in this research? Note that if the LLM is used only for writing, editing, or formatting purposes and does \emph{not} impact the core methodology, scientific rigor, or originality of the research, declaration is not required.
    \item[] Answer: \answerNA{} 
    \item[] Justification: 

\end{enumerate}

\end{document}